\documentclass[11pt]{article}
\usepackage[preprint]{neurips_2024}

\makeatletter
\providecommand{\@LN@col}[1]{}
\providecommand{\@LN}[2]{}
\makeatother

\usepackage[T1]{fontenc}
\usepackage[utf8]{inputenc}
\usepackage[hyphens]{url}
\usepackage{graphicx}
\graphicspath{{files/}{paper/files/}}
\usepackage{caption}
\usepackage{adjustbox}
\usepackage{booktabs}
\usepackage{float}
\usepackage{placeins}
\usepackage{multirow}
\usepackage{amsmath,amssymb}
\usepackage{xspace}
\usepackage{enumitem}
\usepackage{makecell}
\usepackage{amsthm}
\usepackage{microtype}
\usepackage{tabularx}
\usepackage{pifont}
\usepackage{tikz}
\usetikzlibrary{shapes.geometric,calc}
\usepackage{mathtools}
\usepackage{dsfont}

\newcolumntype{C}{>{\centering\arraybackslash}X}
\newtheorem*{remark}{Remark}

\newcommand{\ctbench}{CTBench\xspace}
\newcommand{\netom}{NetO\&M\xspace}
\newcommand{\rca}{RCA\xspace}

\title{CTBench: Evaluating Troubleshooting Capabilities of AI Agents in Realistic Telecom Network Operations}

\author{
Xingyu Yan$^*$, Tingting Dai$^*$, Antonio De Domenico$^{\dagger}$, Mohamed Sana$^{\dagger}$, Nicola Piovesan$^{\dagger}$,\\ \textbf{Changchang Li$^*$, Bowen Liu$^*$, Kun Jiang$^*$, Mengjie Zhang$^*$, Dingcheng Shan$^*$,}\\ \textbf{Jing-Cheng Pang$^*$, Chenwei Wu$^*$, Sijie Wu$^*$, Lianying Chao$^*$, Haoran Cai$^*$,}\\ \textbf{Jiantao Ye$^*$, Xubin Li$^*$, Simon Mark Lucas$^+$, Xin Chen$^*$} \\
$^*${Huawei Technologies, China}\\
$^{\dagger}$Paris Research Center, Huawei Technologies, Boulogne-Billancourt, France\\
$^+${Queen Mary University of London}\\
}

\begin{document}
\maketitle
\begin{abstract}

While autonomous agents show promise for network operations, existing telecom troubleshooting evaluations fail to capture the partial observability and heterogeneity of real-world telecom environments. We introduce \ctbench, a public benchmark evaluating agent troubleshooting capabilities in root cause analysis (RCA) and path restoration. Constructed by domain experts, \ctbench features rich metadata and golden evidence trajectories, utilizing metrics that assess both final accuracy and diagnostic reasoning. Our evaluation of state-of-the-art agents reveals that while they excel at path endpoint identification, they severely underperform in RCA. Crucially, even when agents produce correct answers, they frequently fail to gather the requisite diagnostic evidence demanded in operational practice. Finally, we demonstrate that increased computational expenditure does not guarantee improved diagnostic competence, highlighting critical gaps in current agent capabilities.

\end{abstract}

\section{Introduction}
Network operations and maintenance (\netom) is a demanding operational domain where engineers must interpret heterogeneous telemetry, issue device-specific commands, reconstruct service paths, and diagnose faults across multiple network layers. Unlike many static question-answering settings, \netom requires sequential decision making: an operator must form hypotheses, select diagnostic actions, inspect command outputs, rule out alternative explanations, and produce an evidence-supported conclusion. These requirements make telecom troubleshooting a natural and challenging testbed for evaluating agentic AI systems.

Recent advances in large language model (LLM) agents have demonstrated strong capabilities in multi-step reasoning, tool use, and interactive problem solving~\citep{xi2023rise,wei2022chain,yao2023react}. In communication technology, such agents could assist with root cause analysis, path reconstruction, configuration checking, and closed-loop remediation.
Deployment remains difficult because real networks are partially observable, carry cascading cross-layer faults, and expose multi-vendor command interfaces under strict operational safety requirements.
Existing telecom benchmarks do not reproduce these conditions. They commonly assume full observability or use more homogeneous topologies. \ctbench grounds its tasks in sanitized maintenance cases and CLI outputs recorded from multi-vendor network devices; during evaluation, a tool backend replays those outputs to agents. In addition, these benchmarks cannot distinguish an agent that is unable to localize a fault in a malfunctioning equipment from one that returns a correct label without supporting evidence. For operational \netom, this process-level transparency is a prerequisite for trust, since a diagnosis must be justified before it can be acted on. Furthermore, a realistic benchmark should assess the capabilities of any tool-using agent in a professional setting: acting under partial observability, adapting across heterogeneous interfaces, and attributing network issues across multi-hop propagation rather than stopping at the first visible symptom.

We introduce \ctbench, an agentic benchmark for realistic telecom troubleshooting, covering two operational task families.
In \rca tasks, the agent must identify the affected node and object and assign a normalized root-cause label. In path-restoration tasks, the agent must identify source and destination endpoints and reconstruct the forwarding path.
The benchmark contains 234 expert-curated tasks, 126 for \rca and 108 for path restoration.

Evaluation in \ctbench emphasizes diagnostic reasoning traces rather than merely final accuracy. Authored by 15 senior telecom experts, each task integrates an expert-normalized answer, golden evidence trajectories detailing requisite diagnostic steps, and structural metadata (e.g., observability, heterogeneity, and protocol complexity). To ensure rigorous quality control, independent experts blindly validate all tasks without access to ground-truth data or reference trajectories.

Our contributions are as follows:
\begin{itemize}[leftmargin=*]
    \item We introduce \ctbench, a public benchmark for evaluating agentic troubleshooting in telecom \netom, covering 234 expert-curated \rca and path-restoration tasks.
    \item We define telecom-expert-grounded capability metrics that separately evaluate localization, root-cause identification, path restoration, evidence acquisition, observability robustness, heterogeneous device/vendor handling, and resource usage.
    \item We provide a task metadata schema covering observability, device/vendor heterogeneity, protocol complexity, root-cause count, restored path count, fault-propagation chains, golden solution length, and root-cause categories, enabling fine-grained diagnostic analysis beyond aggregate accuracy.
    \item We evaluate representative agent-model combinations and show that current agents exhibit substantial gaps in evidence-grounded diagnosis, partial-observability reasoning, and robust handling of heterogeneous telecom environments.
\end{itemize}

\section{Related Work}

\subsection{Benchmarks for LLM Agent}

General-purpose LLM agent benchmarks~\citep{mialon2024gaia, jimenez2024swebench, liu2024agentbench, qin2024toolllm, barres2025tau2bench} evaluate multi-step reasoning and tool use across diverse interactive tasks. Process-aware evaluation is also well established: AgentBoard~\citep{ma2024agentboard} measures incremental progress in multi-turn and partially observable environments, while ToolSandbox~\citep{lu2025toolsandbox} evaluates intermediate and final milestones during stateful tool interactions. Building on this work, \ctbench studies telecom troubleshooting through expert-validated diagnostic actions and replayed multi-vendor CLI observations. Its tasks require agents to interpret vendor-specific commands and outputs, reason over network topology and partial device access, and support fault diagnoses with observable evidence. The benchmark therefore measures both final answers and coverage of expert-annotated diagnostic actions in a telecom setting.

\subsection{Benchmarks for Network and Telecom}

Several benchmarks target networking and telecom domains. Knowledge-oriented benchmarks such as TeleQnA \citep{maatouk2025teleqna} and ORAN-Bench \citep{gajjar2025oranbench} evaluate whether models understand telecom specifications or standards. Troubleshooting-oriented benchmarks such as TeleLogs \citep{Sana2025ReasoningLM}, WirelessAgent++ \citep{Tong2026WirelessAgentAA}, and TelcoAgent-Bench \citep{bariah2026telcoagent} move closer to operational network reasoning by introducing diagnostic cases and configuration tasks.

We select NIKA \citep{wang2025networkarenabenchmarkingai}, NetArena \citep{zhou2026netarena}, and NetAgentBench \citep{Twabi2026NetAgentBenchAS} as primary baselines due to their focus on interactive network fault diagnosis in telecom operations (\netom). As summarized in Table~\ref{tab:related_comparison}, \ctbench advances these benchmarks across three critical dimensions:
\textbf{1) Task Types:} While baselines evaluate only Root Cause Analysis (RCA), \ctbench adds 108 path-reconstruction scenarios to its 126 RCA cases, requiring agents to identify endpoints and reconstruct forwarding paths.
\textbf{2) Scenario Features:} Unlike baselines assuming full observability or homogeneous topologies, \ctbench enforces \textit{partial observability} (requiring iterative CLI probing) and \textit{network heterogeneity} (multi-vendor ecosystems like Huawei, Cisco, and H3C).
\textbf{3) Evaluation:} Relative to these telecom benchmarks, \ctbench introduces expert-annotated evidence for trajectory-level evaluation alongside final-answer correctness.

\begin{table}[H]
\centering
\small
\begin{adjustbox}{max width=\textwidth}
\begin{tabularx}{\textwidth}{l CC CC CC}
\toprule
\multirow{2}{*}{\textbf{Benchmark}}
& \multicolumn{2}{c}{\textbf{Task Types}}
& \multicolumn{2}{c}{\textbf{Scenario Features}}
& \multicolumn{2}{c}{\textbf{Evaluation}} \\
\cmidrule(lr){2-3} \cmidrule(lr){4-5} \cmidrule(lr){6-7}
& \textbf{RCA} & \textbf{Path Restoration} & \textbf{Partial Observability} & \textbf{Network Heterogeneity} & \textbf{Expert Evidence} & \textbf{Multi-Dim Metrics} \\
\midrule
NIKA          & $\checkmark$ & $\times$     & $\times$    & $\times$    & $\times$     & $\checkmark$ \\
NetArena      & $\checkmark$ & $\times$     & $\times$    & $\times$    & $\times$     & $\checkmark$ \\
NetAgentBench & $\checkmark$ & $\times$     & $\times$    & $\times$    & $\times$     & $\checkmark$ \\
\midrule
\textbf{\ctbench (ours)} & $\checkmark$ & $\checkmark$ & $\checkmark$& $\checkmark$& $\checkmark$ & $\checkmark$ \\
\bottomrule
\end{tabularx}
\end{adjustbox}
\caption{Conceptual comparison of \ctbench with existing telecom benchmarks across three key dimensions: supported \textbf{task types} (Root Cause Analysis and Path Restoration), \textbf{scenario features} (Partial Observability and Network Heterogeneity), and \textbf{evaluation} depth (Expert-Annotated Evidence and Multi-Dimensional Metrics).}
\label{tab:related_comparison}
\end{table}

\begin{figure*}[t]
\centering
\includegraphics[width=0.95\textwidth]{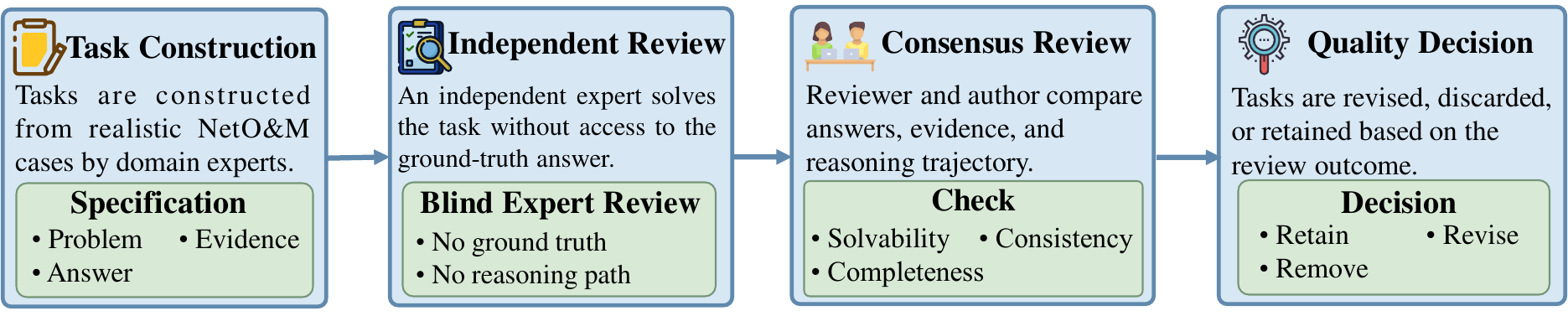}
\caption{Process of expert-involved data quality review for \ctbench. Candidate tasks are sanitized, independently reviewed by telecom experts, checked for answer and evidence consistency, revised when necessary, and retained only after consensus on solvability and operational validity.}
\label{fig:data_quality_workflow}
\end{figure*}

\section{\ctbench Benchmark}

{This section formalizes the tasks in CTBench, and presents the associated datasets.}

\subsection{Task Formulation}

We formulate \ctbench troubleshooting as an interactive decision-making process. Given a task instruction $q_\tau$ ($\tau \in \{\mathrm{RCA}, \mathrm{Path}\}$), an agent $\pi_\theta$ iteratively generates a diagnostic action $a_t \sim \pi_\theta(\cdot \mid q_\tau, h_{t-1})$ conditioned on the history $h_{t-1} = (a_1, o_1, \dots, a_{t-1}, o_{t-1})$. The environment $\mathcal{E}_\tau$ executes $a_t$ to return an observation $o_t = \mathcal{E}_\tau(q_\tau, h_{t-1}, a_t)$ (e.g., CLI outputs, telemetry data, or error states). This iterative interaction yields a complete trajectory $\mathcal{T}(q_\tau) = (q_\tau, a_1, o_1, \dots, a_T, o_T)$, from which the final predicted answer $\hat{y}_\tau$ is decoded.

\paragraph{Root Cause Analysis.} Given a problem description and an interactive diagnostic interface, the agent must iteratively query the environment to formulate a diagnosis. We formalize the target answer as a set of normalized root-cause triples, $y_{\mathrm{RCA}}^* = \{(n_i, o_i, c_i)\}_{i=1}^m$, where $n_i$, $o_i$, and $c_i$ denote the affected node, the specific telecom object (e.g., interface, route, or VPN instance), and the root-cause label, respectively.

\paragraph{Path Restoration.} Agents are required to reconstruct complex service forwarding paths. We formalize the target answer as a set of ordered sequences, $y^*_{\mathrm{Path}} = \{ (p_{i,1}, p_{i,2}, \dots, p_{i,k_i}) \}_{i=1}^{N}$, where each element $p_{i,j}$ denotes a specific routing component or decision (e.g., node, interface, next hop, or policy) within the $i$-th path of length $k_i$.

\subsection{Dataset Construction}
\label{sec:dataset_construction}

\ctbench is constructed via a rigorous expert-in-the-loop pipeline (Figure~\ref{fig:data_quality_workflow}) utilizing sanitized, real-world telecom cases. Fifteen senior experts ($\sim$20 years average experience) formalized all problem statements, interfaces, ground-truth answers, and structural metadata. During evaluation, agents interact with a replay-based mock backend that serves recorded multi-vendor CLI outputs, safely isolating experiments from live network equipment.

To guarantee validity, independent experts blindly solve each candidate task without access to intended trajectories or ground-truth answers. Reviewers subsequently validate the standard answer, annotated golden actions, and all metadata for consistency with available observations. Any disagreements prompt iterative revisions and joint verification to reach author--reviewer consensus. Crucially, candidate tasks are strictly discarded if consensus cannot be reached across four criteria: solvability, answer completeness, golden-action consistency, and metadata correctness.

Through this rigorous filtering, 234 tasks (126 \rca, 108 path-restoration) were ultimately retained from an initial pool of 359 candidates (205 \rca, 154 path-restoration). Notably, 154 of the retained tasks required revision prior to achieving final consensus.

\subsection{Golden Evidence Annotation}
A key feature of \ctbench is that it evaluates not only the final answer $\hat{y}_\tau$, but also the diagnostic process used to obtain it. Each task is associated with an expert-validated golden solution represented as a set of golden actions:
\begin{equation}
    \mathcal{A}^*(q_\tau)
    =
    \{a_1^*,a_2^*,\ldots,a_{T^*}^*\},
    \label{eq:golden_actions}
\end{equation}
where each $a_i^*$ denotes a key action on the expert-defined reference diagnostic route and $T^*$ is the number of annotated actions.

In \ctbench, expert-derived golden actions provide a deterministic reference for evaluating agent operational capabilities. To ensure unambiguous assessment, we exclude alternative equivalent commands during task construction, thereby enforcing a single intended evidence-acquisition route per task (though execution order remains flexible). Consequently, our evidence metrics strictly quantify the coverage of this specific reference route.

Functionally, these golden actions dictate the diagnostic checks requisite for localizing and justifying faults in RCA tasks. For path restoration, they capture the specific checks needed to reconstruct complex forwarding behaviors, such as routing decisions, policy constraints, and tunnel states. Concrete annotation examples are provided in Appendix~\ref{app:task_examples}.

\subsection{Task Metadata}

\ctbench annotates tasks with metadata that enables fine-grained analysis of agent capability.

\paragraph{Evidence Observability.} For RCA tasks, \ctbench defines two levels of evidence observability, summarized in Table~\ref{tab:observability}. This label, denoted by $O(q_\tau)\in\{O_1, O_2\}$, indicates whether a fault evidence can be obtained directly through a permitted diagnostic command, or whether the agent must infer the fault from partial observations obtained through indirect commands.

\begin{table}[!t]
\centering
\small
\setlength{\tabcolsep}{3pt}
\begin{adjustbox}{max width=\textwidth}
\begin{tabularx}{\columnwidth}{@{}cp{0.2\columnwidth}>{\raggedright\arraybackslash}X@{}}
\toprule
Label & Level & Explanation \\
\midrule
O1 & Full & The agent can obtain decisive evidence of the fault by directly implementing a permitted diagnostic command. \\
O2 & Partial & The direct diagnostic command is unavailable or fails. Therefore, the agent must rely on indirect commands that expose only partial observations, requiring additional reasoning and cross-checking to identify the correct root cause. \\

\bottomrule
\end{tabularx}
\end{adjustbox}
\caption{Observability levels in \ctbench.}
\label{tab:observability}
\end{table}

\begin{table}[!t]
\centering
\small
\setlength{\tabcolsep}{3pt}
\begin{adjustbox}{max width=\textwidth}
\begin{tabularx}{\columnwidth}{@{}c>{\raggedright\arraybackslash}Xp{0.18\columnwidth}@{}}
\toprule
Label & Category & Share \\
\midrule
C1 & Interface State and Link-Layer Faults & {28.04\%} \\
C2 & Security, Network Address Translation (NAT), and Edge Access Control & {24.30\%} \\
C3 & Routing Protocol and Policy Control & {16.82\%} \\
C4 & High Availability and Reliability Mechanisms & {15.89\%} \\
C5 & Service, Management, and Other Operational Faults & {14.95\%} \\
\bottomrule
\end{tabularx}
\end{adjustbox}
\caption{RCA category taxonomy and gold-entry distribution.}
\label{tab:rca_categories}
\end{table}

\paragraph{Root-Cause Categories.} Table~\ref{tab:rca_categories} outlines our five-category taxonomy for \rca tasks. Crucially, because a single troubleshooting scenario may encompass multiple concurrent faults, we define these categories at the individual root-cause level rather than the task level, inherently enabling multi-label assignments for complex scenarios.

\paragraph{Root Cause count.} For RCA tasks, \ctbench records the number of independent gold root causes, $N_r(q_\tau)$, to distinguish single-fault scenarios from simultaneous multi-fault failures. Multiple root causes inherently elevate structural difficulty, as agents must disentangle independent faults from cascading downstream symptoms to prevent incomplete or overly broad diagnoses.

\paragraph{Fault-propagation chain.} For RCA tasks, \ctbench annotates the fault-propagation chain length, $N_c(q_\tau)$, which quantifies the causal distance between a latent root cause and its observed symptom. Extended propagation chains inherently elevate diagnostic difficulty, as agents must trace cascading intermediate network states beyond surface-level indicators to isolate the true underlying fault.

\paragraph{Restored path count.} For path-restoration tasks, \ctbench records the number of restored forwarding paths in the gold answer, denoted by $N_{\rm path}(q_\tau)$. A task may require reconstructing a single path or multiple paths caused by branching, redundancy, load balancing, or service-specific forwarding behavior. Multi-path cases are more difficult because the agent must restore path multiplicity rather than only one route.

\paragraph{Protocol complexity.} Across both RCA and path-restoration tasks, \ctbench records protocol complexity, $N_p(q_\tau)$, which quantifies the breadth and interaction depth of underlying mechanisms (e.g., forwarding, control-plane, policy, and redundancy). Elevated complexity dictates that accurate diagnosis requires cross-layer reasoning among interacting protocols, rather than relying on isolated local-state inspections.

\paragraph{Network Heterogeneity.} Network heterogeneity quantifies the operational diversity required to solve a task ($q_\tau$). Let $N_v(q_\tau)$ and $N_d(q_\tau)$ denote the number of distinct vendors and device types strictly involved in the solution trajectory. We define the task-level heterogeneity score as $h_L(q_\tau) = \max(N_v(q_\tau), N_d(q_\tau))$, which we then map to discrete reporting levels using task-specific thresholds for both RCA and path restoration (Table~\ref{tab:heterogeneity}).

\paragraph{Golden Solution Length.}
Across both task types, \ctbench records the golden solution length, $T^*(q_\tau)=\vert{}\mathcal{A}^*(q_\tau)\vert{}$, which quantifies the requisite number of diagnostic steps. This metric formalizes procedural burden: extended golden paths necessitate that agents incrementally aggregate and synthesize broader evidence trails to yield a justified diagnosis.

\begin{table}[!tbp]
\centering
\small
\setlength{\tabcolsep}{4pt}
\begin{adjustbox}{max width=\textwidth}
\begin{tabular}{@{}lp{0.33\columnwidth}ll@{}}
\toprule
Label & Name & RCA & Path Restoration \\
\midrule
H-Low
& Low heterogeneity
&$h_L= 1$
& $h_L\leq 6$ \\

H-High
& High heterogeneity
& $h_L\geq 2$
& $h_L> 6$ \\
\bottomrule
\end{tabular}
\end{adjustbox}
\caption{Network heterogeneity levels based on $h_L=\max(N_v,N_d)$, where $N_v$ and $N_d$ count distinct vendors and device types among devices required to solve the task. RCA and path restoration use different thresholds.}
\label{tab:heterogeneity}
\end{table}

\paragraph{Using Metadata to Characterize Task Difficulty.}
\ctbench leverages these metadata fields to formalize task difficulty as a structural property. Specifically, these dimensions map directly to diagnostic challenges: observability captures information availability; cause/path counts define answer multiplicity; propagation chain length measures causal depth; protocol complexity dictates cross-layer reasoning; vendor heterogeneity reflects operational diversity; and golden solution length indicates requisite expert steps. While RCA difficulty centers on localizing and justifying root causes, path-restoration complexity stems primarily from endpoint identification, forwarding structures, and path multiplicity. Ultimately, we utilize these fine-grained annotations to systematically isolate the operational conditions that most challenge current agents.

\section{Evaluation Protocol}
\label{sec:eval_protoc}
\begin{figure*}[t]
\centering
\includegraphics[width=0.95\textwidth]{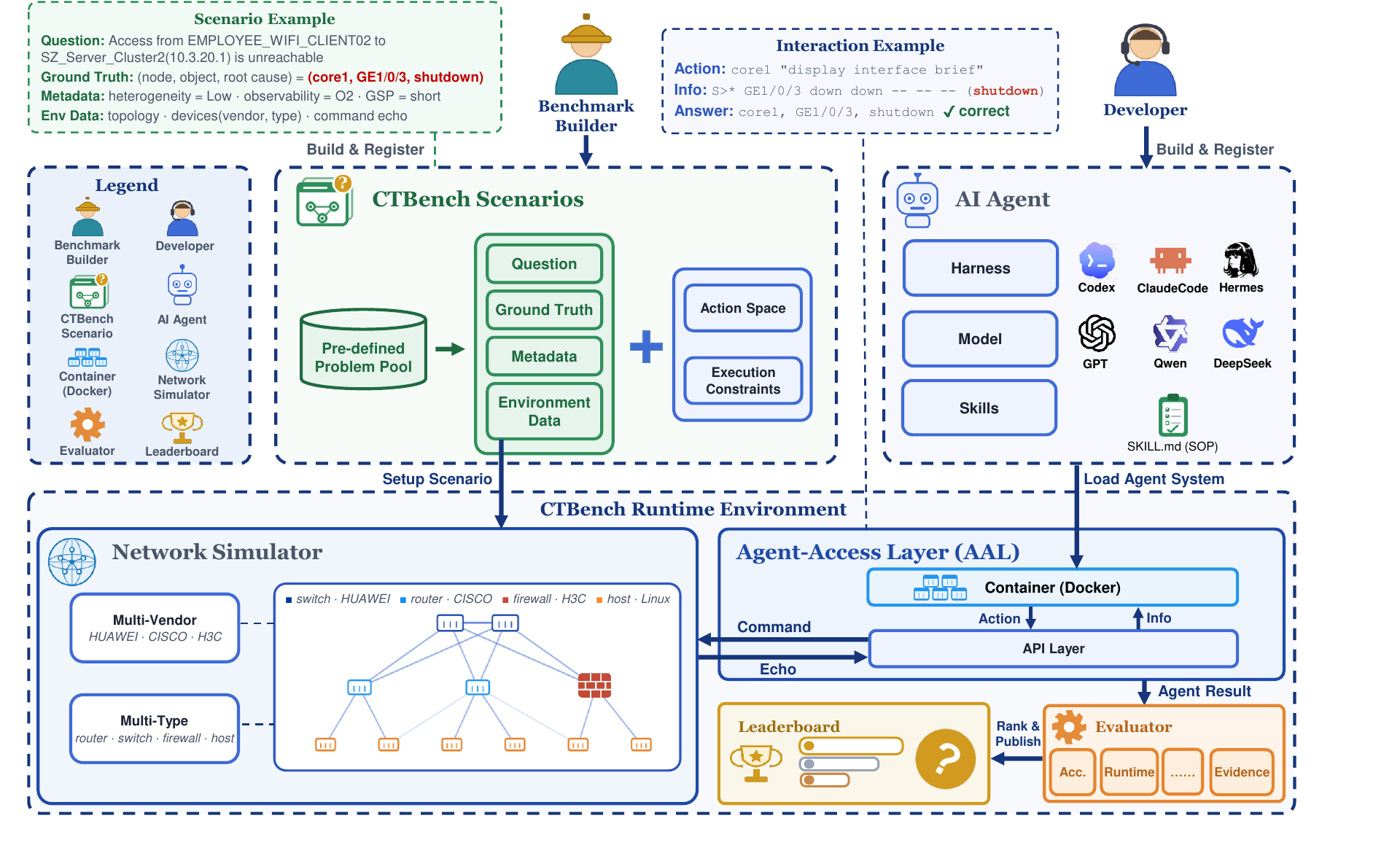}
\caption{\ctbench Automatic Evaluation framework.}
\label{fig:ctbench_evaluation_framework}
\end{figure*}

Figure~\ref{fig:ctbench_evaluation_framework} illustrates the overall evaluation framework of \ctbench. Given a task instance $q_\tau$ and an agent trajectory $\mathcal{T}(q_\tau)$, the evaluator compares the predicted answer $\hat{y}_\tau$ and collected evidence $\hat{\mathcal{A}}(q_\tau)$ against the ground truth $y_\tau^*$ and the related expert-annotated metadata $\mathcal{A}^*(q_\tau)$. In addition to the average task accuracy $\mathrm{Acc}=\mathbb{E}[\mathds{1}(y^*_\tau=\hat{y}_\tau)]$, \ctbench reports task-specific capability
metrics including cost and efficiency metrics.

\subsection{RCA Capability Metrics} \label{sec:rca-capability}
For RCA, \ctbench defines complementary capabilities below.

\paragraph{RCA Localization and Identification.} We evaluate the agent's ability to pinpoint fault locations and classify root causes using Intersection over Union (IoU). Let $L^*$ and $\hat{L}$ denote the gold and predicted sets of affected node-object pairs (e.g., devices and interfaces). Similarly, let $C^*$ and $\hat{C}$ represent the gold and predicted root-cause labels. We compute $\text{RCA-Loc} = \text{IoU}(\hat{L}, L^*)$ and $\text{RCA-ID} = \text{IoU}(\hat{C}, C^*)$ for each task, then report the arithmetic mean of the task-level IoUs.

\paragraph{RCA Evidence.} To quantify coverage of the available expert-validated diagnostic route, we measure the recall of golden actions. Let $\mathcal{A}_{\mathrm{RCA}}^*$ and $\hat{\mathcal{A}}_{\mathrm{RCA}}$ denote the sets of golden evidence steps and executed agent commands, respectively. We compute coverage as $\text{RCA-Evidence} = \frac{\vert{}\mathcal{A}_{\mathrm{RCA}}^* \cap \hat{\mathcal{A}}_{\mathrm{RCA}}\vert{}}{\vert{}\mathcal{A}_{\mathrm{RCA}}^*\vert{}}$. An intersection strictly requires the agent to query the correct device with the exact required command. Across tasks, we pool the covered and total golden-action counts before taking their ratio.

\paragraph{Hierarchical RCA Scoring and Topology-Aware Node Similarity}
To support near-miss analysis, we define hierarchical RCA scoring and topology-aware node similarity. These metrics help distinguish close errors from completely unrelated predictions. The complete taxonomy is given in Appendix~\ref{app:root_cause_hierarchy}; the scoring and similarity procedures are detailed in Appendices~\ref{app:hierarchical_rca} and~\ref{app:node_similarity}.

\subsection{Path Restoration Capability Metrics}
Similar to RCA, \ctbench defines three path-restoration capability metrics below.

\paragraph{Path Localization and Restoration.} We evaluate endpoint identification ($\text{Path-Loc}$) and forwarding path reconstruction ($\text{Path-Rest}$) using Intersection over Union (IoU). Let $P^*_{\text{end}}=\{(\mathrm{src},s^*),(\mathrm{dst},d^*)\}$ and $\hat{P}_{\text{end}}=\{(\mathrm{src},\hat{s}),(\mathrm{dst},\hat{d})\}$ denote the gold and predicted role-labeled endpoints, respectively. Matching an endpoint therefore requires both the correct node and its source or destination role; reversing the two endpoints does not count as a match. Let $P^*$ and $\hat{P}$ represent the reference and predicted directed path-edge sets. We define $\text{Path-Loc} = \text{IoU}(\hat{P}_{\text{end}}, P^*_{\text{end}})$ and $\text{Path-Rest} = \text{IoU}(\hat{P}, P^*)$ for each task, then report the arithmetic mean of the task-level IoUs.

\paragraph{Path Evidence.} To measure coverage of the available expert-validated path-diagnostic route, we compute the recall of golden actions. Let $\mathcal{A}_{\mathrm{Path}}^*$ and $\hat{\mathcal{A}}_{\mathrm{Path}}$ denote the sets of golden evidence steps and executed agent commands, respectively. We compute coverage as $\text{Path-Evidence} = \frac{\vert{}\mathcal{A}_{\mathrm{Path}}^* \cap \hat{\mathcal{A}}_{\mathrm{Path}}\vert{}}{\vert{}\mathcal{A}_{\mathrm{Path}}^*\vert{}}$. Across tasks, we pool the covered and total golden-action counts before taking their ratio.

\subsection{Cost and Efficiency Metrics}

We additionally report latency, interaction rounds, and token consumption. Token consumption is defined as:
\begin{equation}
\text{Tokens}_{\text{total}} =
\text{Tokens}_{\text{input}} + \text{Tokens}_{\text{output}}.
\end{equation}
These quantities are treated as cost and efficiency descriptors rather than substitutes for diagnostic capability.

\begin{table*}[!t]
\centering
\footnotesize
\setlength{\tabcolsep}{2.5pt}
\begin{adjustbox}{max width=\textwidth}
\begin{tabular}{@{}lcccccccc@{}}
\multirow{2}{*}{\textbf{Agent}} & \multicolumn{4}{c}{\textbf{RCA}} & \multicolumn{4}{c}{\textbf{Path Restoration}} \\
\cmidrule(lr){2-5} \cmidrule(lr){6-9}

 & RCA Acc & RCA-Loc & RCA-ID & RCA-Evid. & Path Acc & Path-Loc & Path-Rest & Path-Evid. \\
\midrule
\makecell[l]{CX+GPT-5.5(m)} & \makecell{\textbf{47.62}$\pm$4.5} & \makecell{\textbf{63.99}$\pm$3.7} & \makecell{\textbf{74.74}$\pm$3.7} & \makecell{\textbf{51.64}$\pm$2.5} & \makecell{\textbf{87.96}$\pm$3.2} & \makecell{\textbf{99.38}$\pm$0.6} & \makecell{\textbf{95.42}$\pm$1.4} & \makecell{\textbf{55.81}$\pm$1.2} \\
\makecell[l]{CC+Sonnet-5.5(m)} & \makecell{34.92$\pm$4.3} & \makecell{55.36$\pm$3.5} & \makecell{70.90$\pm$3.0} & \makecell{26.81$\pm$2.3} & \makecell{32.41$\pm$4.5} & \makecell{98.46$\pm$1.1} & \makecell{77.45$\pm$1.9} & \makecell{14.52$\pm$1.2} \\
\makecell[l]{HA+DeepSeek-V4-Pro(h)} & \makecell{17.46$\pm$3.4} & \makecell{37.74$\pm$3.6} & \makecell{49.47$\pm$3.9} & \makecell{41.13$\pm$2.7} & \makecell{26.85$\pm$4.3} & \makecell{85.49$\pm$3.0} & \makecell{65.42$\pm$2.8} & \makecell{21.52$\pm$1.5} \\
\makecell[l]{HA+Qwen3.7-Max(m)} & \makecell{25.40$\pm$3.9} & \makecell{45.03$\pm$3.6} & \makecell{59.46$\pm$3.9} & \makecell{34.95$\pm$2.8} & \makecell{40.74$\pm$4.8} & \makecell{81.48$\pm$3.6} & \makecell{67.48$\pm$3.5} & \makecell{20.25$\pm$1.7} \\
\bottomrule
\end{tabular}
\end{adjustbox}
\caption{Overall pass@1 capabilities (126 RCA, 108 path tasks). Metrics are percentages $\pm$ standard error (SE), reflecting cross-task rather than cross-run variance. Accuracy and IoU denote task means; evidence is pooled golden-action coverage with SE derived from 10,000 bootstrap resamples. Abbreviations: CX=Codex, CC=ClaudeCode, HA=HermesAgent; (m)/(h)=medium/high.}
\label{tab:overall_capability}
\end{table*}

\section{Experiments and Analysis}

We evaluate four agent--LLM configurations: Codex+GPT-5.5, ClaudeCode+Sonnet-5.5, and HermesAgent paired with either DeepSeek-V4-Pro or Qwen3.7-Max.\footnote{Specifically: GPT-5.5(medium), Sonnet-5.5(medium), DeepSeek-V4-Pro(high), and Qwen3.7-Max(medium). Full versioning and release dates are in Appendix~\ref{app:evaluation_environment}.} Agents perform zero-shot inference on a shared task set within an interactive, replay-based telecom sandbox. All metrics are reported as pass@1 (without retries), with standard errors reflecting cross-task—rather than cross-run—variance. We log full execution trajectories alongside efficiency metrics (latency, rounds, tokens), strictly applying uniform answer normalization and evidence-matching protocols across all models.

\subsection{Overall Capability Results}

Table~\ref{tab:overall_capability} details task accuracy and capability metrics. Agents generally perform better on path restoration than RCA, led by Codex+GPT-5.5(medium). Crucially, final accuracy masks severe underlying capability gaps. Even for exactly correct answers, Codex+GPT-5.5(medium) misses complete expert evidence in 47/60 RCA and 93/95 path-restoration tasks, yielding pooled golden-action coverages of merely 60.42\% and 56.15\%, respectively (Table~\ref{tab:correct_answer_evidence}). Similarly, strong path localization (\text{Path-Loc}) does not ensure full restoration (Path IoU) or adequate evidence gathering, as evidenced by HermesAgent+DeepSeek-V4-Pro(high). Furthermore, RCA identification consistently outperforms localization across models (e.g., Codex+GPT-5.5(medium): 74.74\% vs. 63.99\%; HermesAgent+Qwen3.7-Max(medium): 59.46\% vs. 45.03\%), revealing that agents successfully classify fault types but struggle to pinpoint specific faulty nodes.

\subsection{Cost and Efficiency}

\begin{table}[t]
\centering
\footnotesize
\setlength{\tabcolsep}{3pt}
\begin{adjustbox}{max width=\textwidth}
\begin{tabular}{@{}llrrr@{}}
\toprule
Agent & Task & Rds & Lat.(s) & Tok. \\
\midrule
Codex+         & RCA  & \textbf{10.81} & 333.40 & \textbf{476.5k} \\
GPT-5.5(m)     & Path & \textbf{14.14} & 419.80 & \textbf{1019.2k} \\
\addlinespace
ClaudeCode+    & RCA  & 59.34 & \textbf{228.84} & 2706.7k \\
Sonnet-5.5(m)  & Path & 53.08 & \textbf{151.50} & 2162.9k \\
\addlinespace
HermesAgent+   & RCA  & 36.62 & 483.62 & 1491.7k \\
DS-V4-Pro(h)   & Path & 38.66 & 497.84 & 1453.1k \\
\addlinespace
HermesAgent+   & RCA  & 31.09 & 1352.93 & 1218.6k \\
Qwen3.7-M(m)   & Path & 36.77 & 1482.64 & 1574.6k \\
\bottomrule
\end{tabular}
\end{adjustbox}
\caption{Efficiency and cost results. (m)=medium, (h)=high, P=Plus, M=Max, DS=DeepSeek.}
\label{tab:overall_cost}
\end{table}

Table~\ref{tab:overall_cost} details efficiency metrics (interaction rounds, latency, and token usage). Crucially, higher computational cost does not guarantee superior diagnostic capability. For example, ClaudeCode+Sonnet-5.5(medium) consumes significantly more tokens than Codex+GPT-5.5(medium) yet yields lower capability scores. Conversely, HermesAgent+Qwen3.7-Max(medium) outperforms HermesAgent+DeepSeek-V4-Pro(high) in path restoration, but incurs proportional increases in latency and token consumption. These explicit trade-offs underscore the necessity of reporting efficiency metrics independently, rather than treating resource usage as a proxy for agent competence.

\begin{remark}[Human Expert Baseline]
Five domain experts evaluated the final revised task set independently from the task review phase (Figure~\ref{fig:data_quality_workflow}). Each task was answered by one expert, so each expert answered approximately $234/5\approx47$ tasks. Experts answered 102 of 126 RCA tasks correctly (80.95\%) and 104 of 108 path-restoration tasks correctly (96.30\%), requiring 40--60 minutes per task.
\end{remark}

\subsection{RCA Performance by Fault Category}

Figure~\ref{fig:rca_category_accuracy_radar} shows that RCA accuracy varies across the five fault categories (Table~\ref{tab:rca_categories}). C1 accuracy is at most 10.71\% for every evaluated combination, and only Codex+GPT-5.5(medium) obtains nonzero C3 accuracy (40.00\%). C2 has the highest accuracy for Codex+GPT-5.5(medium) and both HermesAgent combinations, whereas ClaudeCode+Sonnet-5.5(medium) performs best on C4. These patterns show that category-level strengths and weaknesses differ across agent--model combinations.

\begin{figure}[!t]
\centering
\begin{tikzpicture}[
    scale=0.72,
    every node/.style={font=\scriptsize},
    grid/.style={gray!35, thin},
    axis/.style={gray!45, thin},
    series/.style={line width=1.2pt, mark size=1.4pt}
]

\def\R{3.0}
\def\rA{0.6}
\def\rB{1.2}
\def\rC{1.8}
\def\rD{2.4}
\def\rE{3.0}

\def\aOne{90}
\def\aTwo{18}
\def\aThree{-54}
\def\aFour{-126}
\def\aFive{162}

\foreach \r in {\rA,\rB,\rC,\rD,\rE} {
    \draw[grid] (0,0) circle (\r);
}
\foreach \a in {\aOne,\aTwo,\aThree,\aFour,\aFive} {
    \draw[axis] (0,0) -- (\a:\R);
}

\node[gray!70, inner sep=1pt] at (-12:\rC) {40\%};
\node[gray!70, inner sep=1pt] at (-12:\rD) {60\%};

\node[align=center, anchor=south, inner sep=2pt] at (\aOne:3.1)
{C1\\Interface/Link};

\node[align=center, anchor=west, inner sep=2pt] at (\aTwo:3.1)
{C2\\Security/NAT\\Access};

\node[align=center, anchor=north west, inner sep=2pt] at (\aThree:3.1)
{C3\\Routing/Policy};

\node[align=center, anchor=north east, inner sep=2pt] at (\aFour:3.1)
{C4\\HA/Reliability};

\node[align=center, anchor=east, inner sep=2pt] at (\aFive:3.1)
{C5\\Service/Mgmt\\Other Ops};

\draw[series, blue, mark=*] plot coordinates {
    (\aOne:0.26775) (\aTwo:2.45) (\aThree:1.5) (\aFour:1.875) (\aFive:1.05) (\aOne:0.26775)
};

\draw[series, red, mark=*] plot coordinates {
    (\aOne:0) (\aTwo:0.841875) (\aThree:0) (\aFour:2.536875) (\aFive:1.171875) (\aOne:0)
};

\draw[series, teal!70!black, mark=*] plot coordinates {
    (\aOne:0.26775) (\aTwo:0.995) (\aThree:0) (\aFour:0.441) (\aFive:0.58575) (\aOne:0.26775)
};

\draw[series, violet, mark=*] plot coordinates {
    (\aOne:0.401625) (\aTwo:1.30) (\aThree:0) (\aFour:0.88) (\aFive:0.82) (\aOne:0.401625)
};

\begin{scope}[shift={(0,-4.0)}]
    \draw[blue, line width=1.2pt, mark=*] (-3.8, 0.3) -- (-3.4, 0.3) node[right, text=black] {Codex+GPT-5.5(m)};
    \draw[violet, line width=1.2pt, mark=*] (1.2, 0.3) -- (1.6, 0.3) node[right, text=black] {HA+Qwen3.7-M(m)};
    \draw[red, line width=1.2pt, mark=*] (-3.8, -0.3) -- (-3.4, -0.3) node[right, text=black] {ClaudeCode+Sonnet-5.5(m)};
    \draw[teal!70!black, line width=1.2pt, mark=*] (1.2, -0.3) -- (1.6, -0.3) node[right, text=black] {HA+DS-V4-Pro(h)};
\end{scope}

\end{tikzpicture}
\caption{RCA accuracy across root-cause categories. Abbreviations: HA=HermesAgent, (m)=medium, (h)=high, P=Plus, M=Max, DS=DeepSeek.}
\label{fig:rca_category_accuracy_radar}
\end{figure}

\subsection{Impact of Observability and Heterogeneity}

We next analyze how task metadata explains agent behavior.

\paragraph{Observability.}
To isolate the impact of indirect evidence, we compare the default O2 setting against an O2$\rightarrow$O1 ablation (providing direct evidence) across 19 RCA tasks (Table~\ref{tab:observability_results}). Reliance on indirect evidence degrades exact accuracy for three configurations: Codex+GPT-5.5(medium) (36.84\% vs. 73.68\%), HermesAgent+DeepSeek-V4-Pro(high) (15.79\% vs. 47.37\%), and HermesAgent+Qwen3.7-Max(medium) (36.84\% vs. 52.63\%). Furthermore, Codex suffers corresponding drops in localization (56.14\% vs. 87.72\%) and identification (45.61\% vs. 82.46\%). Notably, only ClaudeCode+Sonnet-5.5(medium) remains robust, maintaining 73.68\% accuracy across both settings. Finally, because the ablation inherently reduces golden-action requirements (4.32 $\to$ 2.95 average steps, Table~\ref{tab:observability_action_counts}), evidence coverage scores (e.g., Codex: 64.63\% vs. 87.50\%) lack a shared denominator, precluding direct comparative analysis of acquisition behaviors.

\paragraph{Network heterogeneity.}
Table~\ref{tab:heterogeneity_results} demonstrates that high network heterogeneity consistently degrades RCA exact accuracy across all four configurations (e.g., Codex+GPT-5.5(medium): 66.67\% $\to$ 39.08\%; ClaudeCode+Sonnet-5.5(medium): 56.41\% $\to$ 25.29\%). Conversely, path-restoration trends are mixed: exact accuracy drops for Codex+GPT-5.5(medium), HermesAgent+DeepSeek-V4-Pro(high), and HermesAgent+Qwen3.7-Max(medium), yet improves for ClaudeCode+Sonnet-5.5(medium). Path-Rest scores mirror this divergence, decreasing for both HermesAgent configurations, remaining stable for Codex (94.72\% vs. 96.22\%), and increasing for ClaudeCode. Crucially, because the low- and high-heterogeneity subsets comprise distinct tasks, these findings establish correlational associations rather than strict causal effects.

\subsection{Performance on Hierarchical RCA Scoring and Topology-Aware Node Similarity}
The auxiliary exact entry-match and hierarchical scores are reported in Appendix Table~\ref{tab:hierarchical_rca_results}. These recall-oriented measures are computed over gold RCA entries and should not be read as the task-macro IoUs in Table~\ref{tab:overall_capability}. Across the four evaluated agents, hierarchical RCA identification exceeds exact entry-match identification by 0.19--1.12 percentage points, and hierarchical node localization exceeds exact entry-match localization by 9.48--17.14 points. The latter gap suggests that agents sometimes reach the correct topology neighborhood without identifying the exact faulty node or object.


\section{Discussion}

\paragraph{Performance is multi-faceted.}
{A key lesson from \ctbench is that
agent performance should be measured
in detailed ways: separating localization, identification, restoration, and evidence acquisition exposes distinct variations
in abilities.}

\paragraph{Evidence-grounded diagnosis remains difficult.}
Even when agents produce correct final answers, they often fail to gather the full diagnostic evidence identified by telecom experts (Table~\ref{tab:correct_answer_evidence}). This matters for operational trust, where a diagnosis must be justified before remediation.
\paragraph{Partial observability and heterogeneity expose agent limitations.}
Under partial observability, agents must reason from indirect evidence and cross-checks. Under heterogeneity, they must normalize device roles, vendor-specific commands, and output formats. The observed drops in localization, identification, and restoration show that these are not merely implementation details; they are core dimensions of telecom-agent capability.

\section{Conclusion}

We presented \ctbench, an agentic benchmark for realistic telecom network operations and maintenance. Using fine-grained data annotations and multi-dimensional metrics, \ctbench assesses agent proficiency in fault localization, root-cause isolation, path reconstruction, expert-aligned evidence extraction, and reasoning robustness under partial observability and heterogeneous environments. Experiments with representative agent-model combinations reveal that current agents remain far from reliable telecom troubleshooting operators, especially in evidence-grounded \rca and partially observable settings. We hope \ctbench will support future research on trustworthy, domain-grounded, and operationally safe AI agents for communication networks.

\section*{Limitations}

\ctbench covers two troubleshooting task types, path restoration and RCA, and does not evaluate dynamic network repair. Its tool backend replays a finite inventory of recorded CLI outputs, so agents cannot alter device state or respond to live network changes. The reported results therefore do not directly establish performance on live production networks. The benchmark also does not exhaustively cover domains such as full radio-access-network operations, core-network slicing, or cloud-native telecom infrastructure.

\bibliographystyle{plainnat}
\bibliography{reference}

\clearpage
\appendix
\section*{Supplementary Material}

\section{Dataset and Annotation Details}
\label{sec:appendix}
\label{app:dataset_annotation}

\begin{figure*}[t]
\centering
\includegraphics[width=0.95\textwidth]{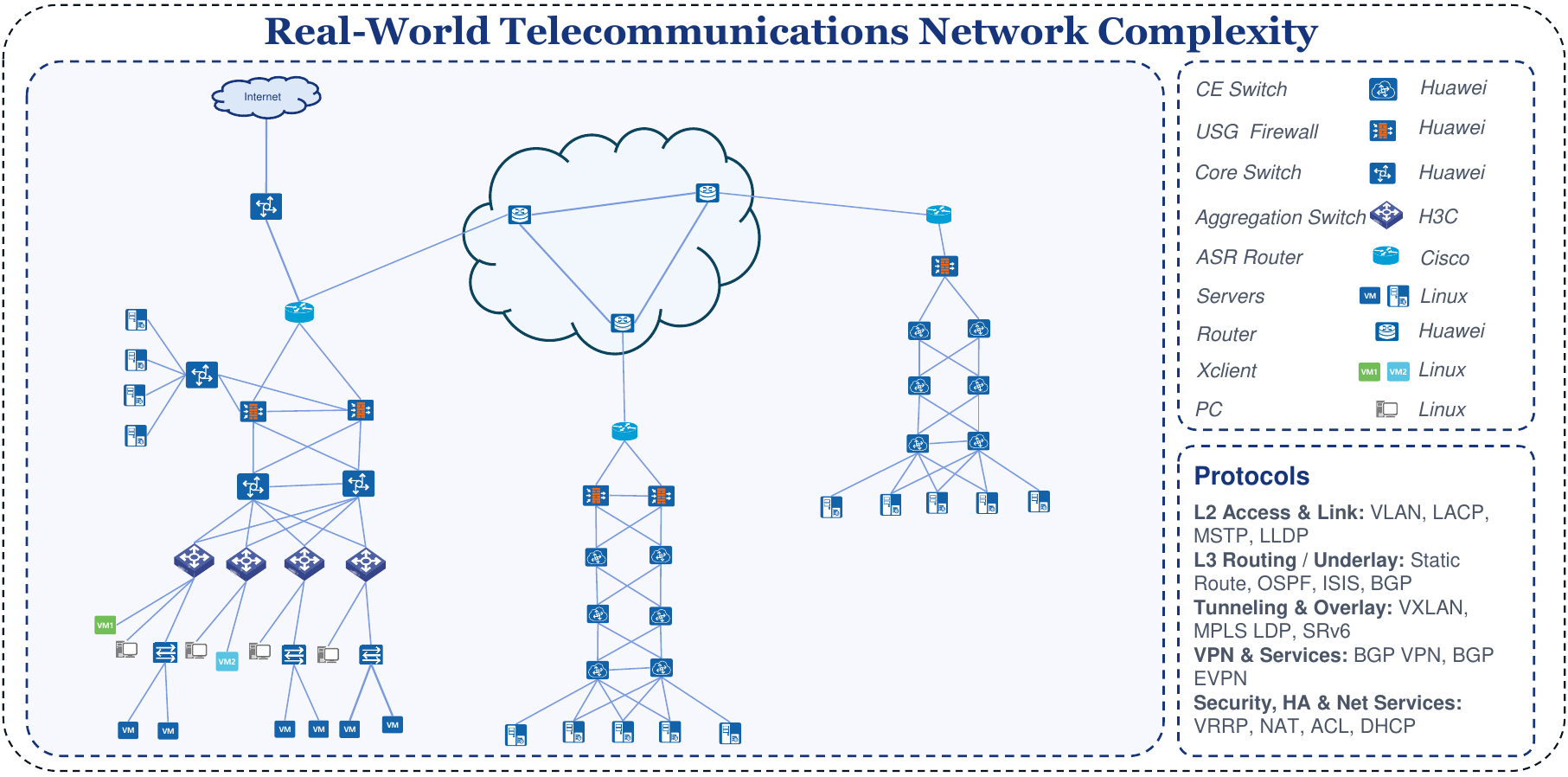}
\caption{Representation of the telecom network under study with multi-vendor equipment and realistic networking protocols.}
\label{fig:real_world_telecom_environments}
\end{figure*}

CTBench is constructed through an expert-in-the-loop process. Its tasks use CLI outputs collected from network equipment spanning multiple vendors, device types, protocols, and interfaces (see Figure \ref{fig:real_world_telecom_environments}). During evaluation, these outputs are replayed through a mock tool backend without live device access. Candidate tasks are abstracted from realistic telecom maintenance cases and sanitized to remove sensitive production information. Fifteen senior telecom domain experts with an average of 20 years of professional experience define the problem statement, available device outputs, permitted command interface, standard answer, and task-side metadata. The current benchmark contains 126 RCA tasks and 108 path-restoration tasks.

\subsection{RCA task}
\label{app:rca_distribution}

The RCA subset contains 126 tasks and 214 normalized root-cause triples. Since a single troubleshooting task may contain multiple independent faults, the number of root-cause entries can exceed the number of tasks. The dataset also contains 176 task-level root-cause mentions, where each root-cause type is counted at most once per task. The difference between 214 entries and 176 mentions reflects repeated affected objects or devices for the same root-cause type within some tasks.

The 17 normalized root-cause labels are grouped into seven operational fault domains for dataset characterization. The main paper merges sparse domains into C5 for stable reporting, while Table~\ref{tab:app_root_cause_category_summary} reports the complete unmerged distribution.

\begin{table*}[t]
\centering
\small
\begin{adjustbox}{max width=\textwidth}
\begin{tabular}{p{0.55\textwidth}rrr}
\toprule
Fault domain & Types & Entries & Ratio \\
\midrule
Interface and link-state faults & 1 & 60 & 28.04\% \\
Security, NAT, and boundary access control & 3 & 52 & 24.30\% \\
Routing protocol and policy control & 2 & 36 & 16.82\% \\
High availability and reliability mechanisms & 2 & 34 & 15.89\% \\
Overlay and VPN service faults & 4 & 26 & 12.15\% \\
Basic addressing and Layer-2 access configuration & 4 & 5 & 2.34\% \\
Monitoring and operational visibility & 1 & 1 & 0.47\% \\
\midrule
Total & 17 & 214 & 100.00\% \\
\bottomrule
\end{tabular}
\end{adjustbox}
\caption{Root-cause category coverage in CTBench fault-localization tasks.}
\label{tab:app_root_cause_category_summary}
\end{table*}

Table~\ref{tab:app_fine_grained_root_causes} reports the fine-grained root-cause distribution. The distribution is intentionally not class-balanced: CTBench preserves frequent operational failure modes while retaining diagnostically distinct long-tail cases. We further merge the fine-grained root-causes into five-category taxonomy shown in Table~\ref{tab:app_merged_rca_category_distribution}.

\begin{table*}[t]
\centering
\small
\setlength{\tabcolsep}{3pt}
\begin{adjustbox}{max width=\textwidth}
\begin{tabular}{p{0.36\textwidth}p{0.30\textwidth}ccc}
\toprule
Fine-grained root cause & Fault domain & Entries & Ratio & Task ment. \\
\midrule
Shutdown & Interface/link state & 60 & 28.04\% & 28 \\
Security policy does not permit the user & Security/NAT/access control & 41 & 19.16\% & 41 \\
IP prefix list misses the corresponding source IP & Routing/policy control & 24 & 11.21\% & 24 \\
Global STP is not enabled & High availability/reliability & 23 & 10.75\% & 23 \\
L3VPN misconfiguration & Overlay/VPN service & 17 & 7.94\% & 17 \\
OSPF misconfiguration & Routing/policy control & 12 & 5.61\% & 6 \\
Global HRP hot-standby protocol is not enabled & High availability/reliability & 11 & 5.14\% & 11 \\
NAT outside-interface attribute error or omission & Security/NAT/access control & 7 & 3.27\% & 7 \\
NAT inside-interface attribute error or omission & Security/NAT/access control & 4 & 1.87\% & 4 \\
SRv6-Policy tunnel planning error & Overlay/VPN service & 4 & 1.87\% & 4 \\
VXLAN misconfiguration & Overlay/VPN service & 4 & 1.87\% & 4 \\
MAC address misconfiguration & Basic L2/access config & 2 & 0.93\% & 2 \\
Interface IP error & Basic L2/access config & 1 & 0.47\% & 1 \\
Interface VLAN misconfiguration & Basic L2/access config & 1 & 0.47\% & 1 \\
VPN configuration missing & Overlay/VPN service & 1 & 0.47\% & 1 \\
Host-information collection missing & Monitoring/visibility & 1 & 0.47\% & 1 \\
Loopback IP conflict & Basic L2/access config & 1 & 0.47\% & 1 \\
\midrule
Total & -- & 214 & 100.00\% & 176 \\
\bottomrule
\end{tabular}
\end{adjustbox}
\caption{Fine-grained root-cause distribution in CTBench. Task mentions count each root-cause type at most once per task.}
\label{tab:app_fine_grained_root_causes}
\end{table*}

\begin{table}[t]
\centering
\small
\setlength{\tabcolsep}{3pt}
\begin{adjustbox}{max width=\textwidth}
\begin{tabular}{lrr}
\toprule
Category & Entries & Ratio \\
\midrule
C1 Interface/link state & 60 & 28.04\% \\
C2 Security/NAT/access & 52 & 24.30\% \\
C3 Routing/policy & 36 & 16.82\% \\
C4 HA/reliability & 34 & 15.89\% \\
C5 Service/management/other & 32 & 14.95\% \\
\bottomrule
\end{tabular}
\end{adjustbox}
\caption{Merged RCA category distribution over 214 gold root-cause entries.}
\label{tab:app_merged_rca_category_distribution}
\end{table}

\subsection{Path Restoration task}

The path-restoration subset contains 108 tasks. Each task asks the agent to reconstruct the forwarding path from a source endpoint to a destination endpoint or destination IP. Unlike RCA tasks, whose answers are sets of root-cause triples, path-restoration answers are ordered network paths. A valid answer must preserve hop order, path multiplicity, and physical egress-interface semantics. The required output uses one line per path, connects consecutive hops with \texttt{->}, writes each non-terminal hop as \texttt{node\_egress-interface}, and writes the terminal endpoint as the node name only.

Each golden path is decomposed into normalized node records with node names, egress interfaces, and raw output segments. This representation supports exact-match scoring while also enabling more diagnostic overlap measures, such as endpoint correctness, ordered edge overlap, interface overlap, and path-count correctness.

\subsection{Annotation Protocol}
\label{app:annotation_protocol}

For RCA tasks, standard answers are normalized into one or more minimal root-cause triples of the form
$(fault\_node, fault\_object, root\_cause)$. For path-restoration tasks, standard answers preserve node order, path multiplicity, and interface semantics. Golden solution paths are annotated as key evidence steps rather than complete command logs. These steps and all task metadata are reviewed through the same independent expert review and author--reviewer consensus process used for task inclusion (Figure~\ref{fig:data_quality_workflow}), including checks of answer and evidence consistency. For multi-root-cause and multi-path tasks, reviewers check answer completeness and evidence sufficiency. The gold action set does not prescribe the order of independent checks. Disagreements concerning the answer, golden evidence steps, or any metadata field are addressed through revision. The task author and reviewing expert jointly verify the revised information and must reach consensus before the task is retained; unresolved cases are excluded.

The review pipeline began with 359 candidate tasks: 205 RCA and 154 path-restoration tasks. After expert review and revision, the final benchmark retained 126 RCA and 108 path-restoration tasks, for 234 tasks in total. The candidate-to-retained yields are 61.46\% for RCA, 70.13\% for path restoration, and 65.18\% overall. Among the 234 retained tasks, 154 (65.81\%) underwent revision before the task author and reviewing expert reached consensus.

\subsection{Metadata Schema}
\label{app:metadata_schema}

CTBench annotates each task with task-side metadata for dataset auditing, structural difficulty analysis, and post-hoc trajectory evaluation. All task metadata undergo the independent expert review and author--reviewer consensus process described in Section~\ref{sec:dataset_construction}. These metadata fields are not provided to agents during inference. For network heterogeneity, vendor and device-type counts include only devices required to solve the task, rather than all devices in the wider network. Table~\ref{tab:app_task_metadata_fields} summarizes the metadata fields used for the two evaluated task types.

\begin{table*}[htb]
\centering
\small
\setlength{\tabcolsep}{3pt}
\renewcommand{\arraystretch}{1.15}
\begin{adjustbox}{max width=\textwidth}
\begin{tabular}{@{}p{0.19\textwidth}p{0.38\textwidth}p{0.38\textwidth}@{}}
\toprule
Metadata name  &Fault localization & Path restoration \\
\midrule
Golden answer &  Normalized root-cause triples in\newline \texttt{standard\_answer.faults} & Ordered endpoint-to-endpoint paths in\newline \texttt{standard\_answer.paths} \\
Root-cause / path count & \texttt{data\_labels.root\_cause\_count};\newline \texttt{data\_labels.has\_multiple\_root\_causes} & \texttt{data\_labels.restored\_path\_count};\newline \texttt{data\_labels.unique\_node\_count} \\
RCA category labels & \texttt{data\_labels.rca\_category\_labels};\newline \texttt{data\_labels.rca\_categories} & Not applicable \\
Evidence observability & \texttt{evidence\_observability.tier};\newline direct/degraded observation pairs & Not applicable \\
Fault-propagation chain & \texttt{fault-propagation\_chain.}\newline \texttt{total\_propagation\_step\_count}; causal propagation steps & Not applicable \\
Golden solution path & \texttt{golden\_solution\_path.}\newline \texttt{golden\_solution\_length}; diagnostic key steps & \texttt{golden\_solution\_path.}\newline \texttt{golden\_solution\_length}; path-reconstruction key steps \\
Network heterogeneity &\texttt{network\_heterogeneity.level};\newline vendor/type counts for $h_L$, device counts and key devices & \texttt{network\_heterogeneity.level};\newline vendor/type counts for $h_L$, device counts and key devices \\
Protocol complexity & \texttt{protocol\_complexity.}\newline \texttt{mechanism\_count}; root-cause, diagnostic, propagation, and observability mechanisms & \texttt{protocol\_complexity.}\newline \texttt{mechanism\_count}; forwarding, diagnostic, and path-restoration mechanisms \\
\bottomrule
\end{tabular}
\end{adjustbox}

\caption{Metadata fields for the two evaluated CTBench task types.}
\label{tab:app_task_metadata_fields}
\end{table*}

\subsection{RCA Metadata Distributions}
\label{app:rca_component_distributions}

Table~\ref{tab:app_rca_question_component_distributions} summarizes the
question-level component distributions for the 126 RCA tasks. These metadata
fields characterize observability, answer multiplicity, causal depth,
evidence-path length, network heterogeneity, and protocol complexity.

\begin{table*}[t]
\centering
\small
\setlength{\tabcolsep}{2pt}
\begin{minipage}[t]{0.32\textwidth}
\centering
\textbf{Evidence Observability}\\
\begin{adjustbox}{max width=\textwidth}
\begin{tabular}{p{0.50\linewidth}rr}
\toprule
Tier & N & Ratio \\
\midrule
O1 direct & 107 & 84.92\% \\
O2 indirect & 19 & 15.08\% \\
\bottomrule
\end{tabular}
\end{adjustbox}
\end{minipage}
\hfill
\begin{minipage}[t]{0.32\textwidth}
\centering
\textbf{Root-Cause Count}\\
\begin{tabular}{p{0.50\linewidth}rr}
\toprule
Tier & N & Ratio \\
\midrule
RC1 & 66 & 52.38\% \\
RC2 & 44 & 34.92\% \\
RC3+ & 16 & 12.70\% \\
\bottomrule
\end{tabular}
\end{minipage}
\hfill
\begin{minipage}[t]{0.32\textwidth}
\centering
\textbf{Fault-Propagation Chain}\\
\begin{tabular}{p{0.50\linewidth}rr}
\toprule
Tier & N & Ratio \\
\midrule
Low ($\leq4$) & 19 & 15.08\% \\
Medium (5--13) & 87 & 69.05\% \\
High ($>13$) & 20 & 15.87\% \\
\bottomrule
\end{tabular}
\end{minipage}

\begin{minipage}[t]{0.32\textwidth}
\centering
\textbf{Golden Solution Path}\\
\begin{tabular}{p{0.50\linewidth}rr}
\toprule
Tier & N & Ratio \\
\midrule
Short ($\leq4$) & 73 & 57.94\% \\
Long ($>4$) & 53 & 42.06\% \\
\bottomrule
\end{tabular}
\end{minipage}
\hfill
\begin{minipage}[t]{0.32\textwidth}
\centering
\textbf{Device Heterogeneity}\\
\begin{tabular}{p{0.50\linewidth}rr}
\toprule
Tier & N & Ratio \\
\midrule
H-Low (=1) & 39 & 30.95\% \\
H-High (>1) & 87 & 69.05\% \\
\bottomrule
\end{tabular}
\end{minipage}
\hfill
\begin{minipage}[t]{0.32\textwidth}
\centering
\textbf{Protocol Complexity}\\
\begin{tabular}{p{0.50\linewidth}rr}
\toprule
Tier & N & Ratio \\
\midrule
Low/Mod. ($\leq7$) & 76 & 60.32\% \\
High ($>7$) & 50 & 39.68\% \\
\bottomrule
\end{tabular}
\end{minipage}
\caption{RCA question-level component distributions over 126 tasks.}
\label{tab:app_rca_question_component_distributions}

\end{table*}

The O2 subset contains tasks in which decisive evidence must be inferred from indirect observations. The multi-root-cause subset tests answer completeness. Fault-propagation-chain length captures causal depth, while golden solution length captures the number of expert-required evidence steps. Device
heterogeneity and protocol complexity capture the operational diversity and cross-layer reasoning burden of the task.

\subsection{Path-Restoration Metadata Distributions}
\label{app:path_component_distributions}

The path-restoration subset contains 108 tasks. Table~\ref{tab:app_path_question_component_distributions} summarizes the main path-restoration metadata distributions. Component counts are computed from normalized gold paths and metadata in the released task files.

\begin{table*}[t]
\centering

\small
\setlength{\tabcolsep}{3pt}
\begin{minipage}[t]{0.48\textwidth}
\centering
\textbf{Restored Path Count}\\
\begin{adjustbox}{max width=\textwidth}
\begin{tabular}{p{0.56\linewidth}rr}
\toprule
Tier & N & Ratio \\
\midrule
P1 one path & 64 & 59.26\% \\
P2 two paths & 20 & 18.52\% \\
P3+ three or more & 24 & 22.22\% \\
\bottomrule
\end{tabular}
\end{adjustbox}
\end{minipage}
\hfill
\begin{minipage}[t]{0.48\textwidth}
\centering
\textbf{Golden Solution Path}\\
\begin{tabular}{p{0.56\linewidth}rr}
\toprule
Tier & N & Ratio \\
\midrule
Short/Mod. ($\leq33$) & 74 & 68.52\% \\
Long ($>33$) & 34 & 31.48\% \\
\bottomrule
\end{tabular}
\end{minipage}

\begin{minipage}[t]{0.48\textwidth}
\centering
\textbf{Device Heterogeneity}\\
\begin{tabular}{p{0.56\linewidth}rr}
\toprule
Tier & N & Ratio \\
\midrule
H-Low ($\leq6$) & 58 & 53.70\% \\
H-High ($>6$) & 50 & 46.30\% \\
\bottomrule
\end{tabular}
\end{minipage}
\hfill
\begin{minipage}[t]{0.48\textwidth}
\centering
\textbf{Protocol Complexity}\\[-0.3ex]
\begin{tabular}{p{0.56\linewidth}rr}
\toprule
Tier & N & Ratio \\
\midrule
Low/Mod. ($\leq4$) & 48 & 44.44\% \\
High ($>4$) & 60 & 55.56\% \\
\bottomrule
\end{tabular}
\end{minipage}
\caption{Path-restoration question-level component distributions over 108 tasks.}
\label{tab:app_path_question_component_distributions}
\end{table*}

Most path-restoration tasks have one gold path, but 44 tasks require multiple paths. These multi-path cases stress branching, redundancy, load balancing, and service-specific forwarding behavior. Path-restoration tasks also have longer evidence paths than RCA tasks because the agent must reconstruct
complete forwarding behavior rather than identify one localized root cause.

\subsection{Task Examples}
\label{app:task_examples}

We provide one RCA example and one path-restoration example to illustrate how
task descriptions, gold answers, metadata, and golden evidence steps are
connected. Agents receive only the task description and permitted tools during
evaluation; they do not receive gold answers, metadata labels, or golden
evidence steps.

\paragraph{RCA example.}
RCA-90 asks for the minimal set of root causes explaining why ping from \texttt{Site1 area, Access-PC-01} to \texttt{20.1.1.10} is unreachable. The gold answer contains two port-fault entries: \texttt{BoardLeaf-01;GE1/0/0;shutdown} and \texttt{BoardLeaf-01;GE1/0/1;shutdown}. Both entries belong to C1 interface-state faults. The task is an RC2 task with two independent gold root causes. It is O1 because decisive interface-state evidence is directly observable on \texttt{BoardLeaf-01}. The golden solution path has three key steps, including interface description, current configuration, and LLDP neighbor checks. Its heterogeneity score is $h_L=1$ under the distinct-vendor and distinct-device-type definition, placing it in H-Low.

\paragraph{Path-restoration example.}
PR-100 asks the agent to restore the path from \texttt{SZ\_Server\_Cluster3} in the Shenzhen data center to \texttt{SH\_SAL\_PC01} at \texttt{10.2.10.1} in the Shanghai branch. The gold
answer contains one 12-element path from \texttt{SZ\_Server\_Cluster3} through the Shenzhen core, PE devices, Beijing gateway, Shanghai aggregation, and finally \texttt{SH\_SAL\_PC01}. The task is a P1 task because the normalized gold answer contains one path. Its golden solution path has 33 expert key steps, placing it at the upper boundary of the Short/Moderate tier.
Representative evidence includes route-table checks and LLDP-neighbor checks.
Its heterogeneity score is $h_L=6$ under the same distinct-vendor and distinct-device-type definition, placing it in Path H-Low ($h_L\leq6$); its protocol-complexity mechanism count is four.

\section{Supplementary Evaluation Metrics}
\label{app:supp_metrics}

The main paper reports component-level IoUs for RCA localization and identification and for path localization and restoration. This section defines supplementary metrics used only for diagnostic error analysis. These metrics do not replace the primary scores.

\subsection{Structural Consistency Metrics}
\label{app:structural_consistency}

The primary RCA localization and identification IoUs score node--object pairs and cause labels separately. To retain their assignment, we treat each normalized $(\text{node},\text{object},\text{cause})$ triple as one set element and compute its set IoU for each RCA task. For path restoration, the primary directed-edge IoU does not retain which edges belong to the same path. We therefore treat each complete ordered path as one set element and compute path-set IoU for each path task. Paths are normalized with the same parser used for the edge metric, including the egress interface encoded in each path token when present. Both scores are arithmetic means of the per-task IoUs over 126 RCA or 108 path-restoration tasks, respectively. An unanswered or unparsable prediction yields an empty predicted set and a score of zero because each gold set is nonempty.

\begin{table*}[t]
\centering
\small
\setlength{\tabcolsep}{6pt}
\begin{adjustbox}{max width=\textwidth}
\begin{tabular}{@{}lrr@{}}
\toprule
Agent--model configuration & RCA Triplet Macro IoU & Path Sequence Macro IoU \\
\midrule
Codex+GPT-5.5(m) & 62.08\% & 87.96\% \\
ClaudeCode+Sonnet-5.5(m) & 54.56\% & 37.38\% \\
HermesAgent+DeepSeek-V4-Pro(h) & 31.53\% & 27.31\% \\
HermesAgent+Qwen3.7-Max(m) & 42.71\% & 42.59\% \\
\bottomrule
\end{tabular}
\end{adjustbox}
\caption{Supplementary structural-consistency scores for the four configurations in Table~\ref{tab:overall_capability}. Each score is a per-task macro average. RCA triplet IoU preserves the binding between location and cause; path sequence IoU preserves complete ordered paths rather than only their pooled directed edges. (m)=medium, (h)=high.}
\label{tab:structural_consistency}
\end{table*}

\subsection{Hierarchical RCA Scoring}
\label{app:hierarchical_rca}

Strict RCA identification treats all incorrect root-cause labels equally. This binary treatment hides whether an incorrect label is outside the relevant mechanism family or only differs at a finer taxonomy. We therefore report a supplementary hierarchical RCA score based on the expert-defined RCA taxonomy in Table~\ref{tab:hierarchical_rca_taxonomy}. The score gives partial credit when the predicted root-cause label matches the gold label at coarser levels of the taxonomy, such as the same subtype, technical domain, or broad fault family. This follows the common use of semantic hierarchies in classification analysis, for example in ImageNet-style label taxonomies~\cite{deng2009imagenet,russakovsky2015imagenet}.

We use this score only for diagnostic analysis of failed trajectories. It separates predictions that are unrelated to the gold root cause from predictions that identify a related mechanism but choose the wrong normalized label. The primary benchmark metrics remain exact task accuracy and task-macro RCA IoU; hierarchical RCA is an auxiliary measure for interpreting near misses.

Let $\mathrm{sim}(c,\hat{c})$ denote the similarity between a gold root-cause label $c$ and a predicted label $\hat{c}$. We assign similarity $1.0$ for an exact label match, $0.8$ for the same fine-grained subtype, $0.6$ for the same technical domain, $0.4$ for the same super-category, and $0$ otherwise. For multi-root-cause tasks, we define $G_q$ as the set of gold RCA entries and $P_q$ as the set of predicted RCA entries, compute all pairwise similarities between predicted and gold RCA entries, perform maximum-weight bipartite matching, and normalize the matched similarity by the number of gold RCA entries:
\begin{equation}
\begin{gathered}
\text{Hier-RCA}(q) = 
\displaystyle\frac{
\max_{M \in \mathcal{M}(P_q,G_q)}
\sum_{(p,g)\in M}
\mathrm{sim}(c_g,c_p)
}
{|G_q|}.
\end{gathered}
\end{equation}

\begin{table*}[htb]
\centering

\small
\begin{adjustbox}{max width=\textwidth}
\begin{tabular}{p{0.18\textwidth}p{0.18\textwidth}p{0.54\textwidth}}
\toprule
Match level & Similarity & Interpretation \\
\midrule
Exact label & 1.0 & The predicted normalized root-cause label is identical to the gold label. \\
Same subtype & 0.8 & The prediction falls under the same fine-grained failure subtype but misses the exact normalized label. \\
Same technical domain & 0.6 & The prediction belongs to the same operational domain, such as security/NAT access control, routing/policy control, or high-availability mechanisms. \\
Same super-category & 0.4 & The prediction remains within the same broad fault family, such as connectivity, policy, service, redundancy, or operational-visibility faults. \\
Unrelated & 0.0 & The prediction belongs to a different fault family or cannot be mapped to the root-cause taxonomy. \\
\bottomrule
\end{tabular}
\end{adjustbox}
\caption{Root-cause hierarchy similarity used by hierarchical RCA scoring.}
\label{tab:hierarchical_rca_taxonomy}
\end{table*}

\subsection{Topology-Aware Node Similarity}
\label{app:node_similarity}

We also define a topology-aware node similarity score as a supplement to exact node-object localization. Exact node matches receive full credit. Non-identical nodes receive partial credit when they belong to the same logical cluster, same site and role, or nearby topology neighborhood (see Table~\ref{tab:hierarchical_node_similarity}). This score evaluates node proximity only; it does not include object-level or root-cause-label similarity.

\begin{table*}[htb]
\centering
\small
\begin{adjustbox}{max width=\textwidth}
\begin{tabular}{p{0.22\textwidth}p{0.18\textwidth}p{0.50\textwidth}}
\toprule
Node relation & Similarity component & Interpretation \\
\midrule
Exact node & 1.0 & The normalized predicted node is identical to the gold node. \\
Same logical cluster & 0.7 taxonomy score & The nodes are in the same redundancy pair, HA pair, or inferred logical device cluster. \\
Same site and role & 0.5 taxonomy score & The nodes share both site and device-role metadata. \\
Same site or same role & 0.3 taxonomy score & The nodes share only one of site or role. \\
Topology distance 1 & 0.6 topology score & The nodes are one hop apart in the LLDP-derived topology graph. \\
Topology distance 2 & 0.3 topology score & The nodes are two hops apart in the LLDP-derived topology graph. \\
Disconnected or farther & 0.0 topology score & The nodes are disconnected, unknown, or at distance at least three. \\
\bottomrule
\end{tabular}
\end{adjustbox}
\caption{Topology-aware node similarity used by hierarchical node location.}
\label{tab:hierarchical_node_similarity}
\end{table*}

For non-identical nodes, the final hierarchical node similarity combines the taxonomy and topology components:
\begin{equation}
\begin{aligned}
\mathrm{node\_sim} ={}& 0.7 \cdot \mathrm{taxonomy\_sim} + 0.3 \cdot \mathrm{topology\_sim}.
\end{aligned}
\end{equation}
If either node is absent from the LLDP-derived topology, the score falls back to the taxonomy component inferred from normalized node names, site, role, and redundancy metadata.

\begin{table}[t]
\centering
\footnotesize
\setlength{\tabcolsep}{4pt}
\begin{adjustbox}{max width=\textwidth}
\begin{tabular}{@{} l cccc @{}}
\toprule
\multirow{2}{*}{Agent} & \multicolumn{2}{c}{RCA-ID} & \multicolumn{2}{c}{RCA-Loc} \\
\cmidrule(lr){2-3} \cmidrule(l){4-5}
& Exact & Hier. & Exact & Hier. \\
\midrule
\begin{tabular}[c]{@{}l@{}}Codex+\\GPT-5.5(medium)\end{tabular}       & 71.96 & 72.52 & 64.02 & 73.50 \\ \addlinespace
\begin{tabular}[c]{@{}l@{}}ClaudeCode+\\Sonnet-5.5(medium)\end{tabular}     & 59.81 & 60.00 & 50.93 & 61.29 \\ \addlinespace
\begin{tabular}[c]{@{}l@{}}HermesAgent+\\DeepSeek-V4-Pro(high)\end{tabular} & 49.07 & 50.19 & 41.12 & 58.26 \\ \addlinespace
\begin{tabular}[c]{@{}l@{}}HermesAgent+\\Qwen3.7-Max(medium)\end{tabular}     & 61.68 & 62.06 & 49.53 & 63.84 \\ \addlinespace
\bottomrule
\end{tabular}
\end{adjustbox}
\caption{Exact entry-match and hierarchical RCA scores (\%). Both are auxiliary, recall-oriented scores over gold RCA entries, distinct from the task-macro IoUs in Table~\ref{tab:overall_capability}.}
\label{tab:hierarchical_rca_results}
\end{table}

\subsection{Root-Cause Type Hierarchy}
\label{app:root_cause_hierarchy}

Table~\ref{tab:rca_root_cause_type_hierarchy} lists the normalized root-cause labels defined by the \rca output schema and their hierarchical grouping. The L0 column contains the complete set of schema-level root-cause labels that a model may output under the task instructions, including labels that do not appear as gold answers in the current 126-task release. L1, L2, and L3 provide progressively coarser groupings for near-miss analysis. Legacy gold-answer aliases are normalized to these schema-level L0 labels before hierarchical matching.

\begin{table*}[t]
\centering
\small
\setlength{\tabcolsep}{3pt}
\begin{adjustbox}{max width=\textwidth}
\begin{tabular}{p{0.18\textwidth}p{0.20\textwidth}p{0.23\textwidth}p{0.29\textwidth}}
\toprule
L3 super-category & L2 technical domain & L1 root-cause subtype & L0 normalized root-cause label \\
\midrule
Physical / Link State & Interface state & Administrative shutdown & shutdown \\
Physical / Link State & Interface performance & Bandwidth congestion & traffic congestion occupying port bandwidth \\
Physical / Link State & Interface MTU & MTU mismatch & MTU value misconfiguration \\
Security / Boundary Control & Security policy & Missing permit rule & security policy rule does not permit the corresponding user \\
Security / Boundary Control & NAT role binding & NAT external-interface attribute error & NAT external interface attribute misconfiguration or missing configuration \\
Security / Boundary Control & NAT role binding & NAT internal-interface attribute error & NAT internal interface attribute misconfiguration or missing configuration \\
Routing / Policy Control & Static route & Blackhole route & blackhole route \\
Routing / Policy Control & Static route & Missing static route & missing static route \\
Routing / Policy Control & Static route & Incorrect static route & incorrect static route \\
Routing / Policy Control & ARP / adjacency & ARP configuration & ARP configuration error \\
Routing / Policy Control & Forwarding loop & Layer-3 loop & Layer 3 loop \\
Routing / Policy Control & EGP control plane & BGP configuration & BGP configuration error \\
Routing / Policy Control & Route policy & Prefix-list coverage missing & IP prefix list missing the corresponding user source IP address \\
Routing / Policy Control & IGP control plane & OSPF configuration & OSPF configuration error \\
Routing / Policy Control & IGP control plane & ISIS configuration & ISIS configuration error \\
HA / Reliability & Layer-2 loop prevention & STP not enabled & global STP not enabled \\
HA / Reliability & Layer-2 loop prevention & Port STP not enabled & port STP not enabled \\
HA / Reliability & Hot-standby redundancy & VRRP/HRP redundancy not enabled & global VRRP hot standby redundancy protocol not enabled \\
Service / Overlay & VPN service isolation & L3VPN configuration & L3VPN configuration error \\
Service / Overlay & VPN service isolation & L2VPN configuration & L2VPN configuration error \\
Service / Overlay & VXLAN overlay & VXLAN configuration & VXLAN configuration error \\
Service / Overlay & SRv6 policy tunnel & SRv6-Policy planning & SRV6-Policy tunnel planning error \\
Service / Overlay & VPN service provisioning & Missing VPN configuration & VPN configuration missing \\
Access Configuration & Layer-2 forwarding identity & MAC address configuration & MAC address configuration error \\
Access Configuration & Interface addressing & Interface IP configuration & interface IP error \\
Access Configuration & VLAN access binding & Interface VLAN configuration & interface VLAN configuration error \\
Access Configuration & Address uniqueness & Loopback IP conflict & loopback IP configuration conflict \\
Operations Support & Host information collection & Missing host-information collection & host information collection function missing \\
\bottomrule
\end{tabular}
\end{adjustbox}

\caption{Root-cause type hierarchy used for hierarchical RCA scoring over the complete RCA output-schema label space.}
\label{tab:rca_root_cause_type_hierarchy}
\end{table*}

\FloatBarrier

\section{Supplementary Evaluation Results}
\label{app:supp_results}

This section provides detailed result tables omitted from the main paper.

\subsection{Evidence Coverage on Correct Answers}
\label{app:correct_answer_evidence}

Table~\ref{tab:correct_answer_evidence} restricts the evidence analysis to questions with exactly correct final answers. A question is incomplete when at least one expert-annotated golden action was not executed. The frequency is the number of such questions divided by the number of exactly correct questions; evidence coverage instead pools covered and total golden-action counts over those questions.

\begin{table*}[t]
\centering
\small
\setlength{\tabcolsep}{5pt}
\begin{adjustbox}{max width=\textwidth}
\begin{tabular}{llrrrrr}
\toprule
Agent & Task & Correct & Incomplete & Freq. (\%) & Covered/Gold & Cov. (\%) \\
\midrule
Codex+GPT-5.5(m) & RCA & 60 & 47 & 78.33 & 229/379 & 60.42 \\
Codex+GPT-5.5(m) & Path & 95 & 93 & 97.89 & 2263/4030 & 56.15 \\
ClaudeCode+Sonnet-5.5(m) & RCA & 44 & 44 & 100.00 & 84/299 & 28.09 \\
ClaudeCode+Sonnet-5.5(m) & Path & 35 & 35 & 100.00 & 290/1390 & 20.86 \\
HermesAgent+DeepSeek-V4-Pro(h) & RCA & 22 & 19 & 86.36 & 68/121 & 56.20 \\
HermesAgent+DeepSeek-V4-Pro(h) & Path & 29 & 29 & 100.00 & 242/918 & 26.36 \\
HermesAgent+Qwen3.7-Max(m) & RCA & 32 & 30 & 93.75 & 72/179 & 40.22 \\
HermesAgent+Qwen3.7-Max(m) & Path & 44 & 44 & 100.00 & 375/1326 & 28.28 \\
\bottomrule
\end{tabular}
\end{adjustbox}
\caption{Golden-action coverage restricted to exactly correct answers. Incomplete counts questions with per-question coverage below 100\%; Freq. divides Incomplete by Correct. Cov. pools covered and total golden actions over correct questions. Matching requires the correct device and command; the metric measures reference-route coverage, not whether an observation informed the final answer. (m)=medium, (h)=high.}
\label{tab:correct_answer_evidence}
\end{table*}

\subsection{Full RCA Category Results}
\label{app:category_results}

Table~\ref{tab:category_results} reports localization, identification, and evidence scores for each model across the five merged RCA categories. These results complement the RCA category radar plot in the main paper.

\begin{table*}[t]
\centering
\begin{adjustbox}{max width=\textwidth}
\begin{tabular}{llcccc}
\toprule
Agent & Category & RCA-Acc & RCA-Loc & RCA-ID & RCA-Evid. \\
\midrule
Codex+GPT-5.5(m) & C1 & 7.14 & 51.19 & 78.57 & 32.22 \\
Codex+GPT-5.5(m) & C2 & 65.31 & 98.64 & 97.96 & 71.11 \\
Codex+GPT-5.5(m) & C3 & 40.00 & 40.00 & 80.00 & 59.52\\
Codex+GPT-5.5(m) & C4 & 50.00 & 52.94 & 52.94 & 54.93 \\
Codex+GPT-5.5(m) & C5 & 28.12 & 62.50 & 71.88 & 42.98 \\
\midrule
ClaudeCode+Sonnet-5.5(m) & C1 & 0.00 & 55.95 & 92.86 & 24.44 \\
ClaudeCode+Sonnet-5.5(m) & C2 & 22.45 & 37.76 & 45.92 & 24.44 \\
ClaudeCode+Sonnet-5.5(m) & C3 & 0.00 & 38.33 & 93.33 & 47.62 \\
ClaudeCode+Sonnet-5.5(m) & C4 & 67.65 & 67.65 & 67.65 & 25.33 \\
ClaudeCode+Sonnet-5.5(m) & C5 & 31.25 & 43.75 & 46.88 & 19.83 \\
\midrule
HermesAgent+DeepSeek-V4-Pro(h) & C1 & 7.14 & 40.48 & 67.86 & 31.11 \\
HermesAgent+DeepSeek-V4-Pro(h) & C2 & 26.53 & 65.99 & 70.41 & 60.74 \\
HermesAgent+DeepSeek-V4-Pro(h) & C3 & 0.00 & 0.00 & 76.67 & 60.71 \\
HermesAgent+DeepSeek-V4-Pro(h) & C4 & 11.76 & 14.71 & 14.71 & 34.21 \\
HermesAgent+DeepSeek-V4-Pro(h) & C5 & 15.62 & 39.06 & 42.19 & 39.67 \\
\midrule
HermesAgent+Qwen3.7-Max(m) & C1 & 10.71 & 50.60 & 67.86 & 23.33 \\
HermesAgent+Qwen3.7-Max(m) & C2 & 34.69 & 91.50 & 95.92 & 68.89 \\
HermesAgent+Qwen3.7-Max(m) & C3 & 0.00 & 0.00 & 65.00 & 50.00 \\
HermesAgent+Qwen3.7-Max(m) & C4 & 23.53 & 23.53 & 35.29 & 26.32 \\
HermesAgent+Qwen3.7-Max(m) & C5 & 21.88 & 34.38 & 31.25 & 22.31 \\
\bottomrule
\end{tabular}
\end{adjustbox}

\caption{RCA results by fault category. Localization and identification are macro IoUs across task--category memberships; a task may contribute to multiple categories. Evidence is golden-action coverage. Scores are percentages. (m)=medium, (h)=high.}
\label{tab:category_results}
\end{table*}

\subsection{Observability Results for All Agents}
\label{app:observability_results}

Table~\ref{tab:observability_results} reports RCA performance by observability level for all evaluated agents. The O2 subset contains tasks in which decisive evidence is unavailable or must be inferred from indirect observations. The O2$\rightarrow$O1 rows correspond to an ablation setting where the originally unobservable elements are made directly observable. Table~\ref{tab:observability_action_counts} reports the corresponding golden-action and agent-command counts. Since the ablation regenerates the golden evidence specification, its evidence scores describe overlap with a different action set and are not used to infer an evidence-acquisition improvement over O2.

\begin{table*}[t]
\centering
\small
\setlength{\tabcolsep}{4pt}
\begin{adjustbox}{max width=\textwidth}
\begin{tabular}{lcccccc}
\toprule
Agent & Obs. & N & Acc. & RCA-Loc & RCA-ID & RCA-Evid. \\
\midrule
Codex+GPT-5.5(m) & O1 & 107 & 49.53 & 65.39 & 79.91 & 50.07 \\
Codex+GPT-5.5(m) & O2 & 19 & 36.84 & 56.14 & 45.61 & 64.63 \\
Codex+GPT-5.5(m) & O2$\rightarrow$O1 & 19 & 73.68 & 87.72 & 82.46 & 87.50 \\
ClaudeCode+Sonnet-5.5(m) & O1 & 107 & 28.04 & 49.92 & 69.16 & 26.95 \\
ClaudeCode+Sonnet-5.5(m) & O2 & 19 & 73.68 & 85.96 & 80.70 & 25.61 \\
ClaudeCode+Sonnet-5.5(m) & O2$\rightarrow$O1 & 19 & 73.68 & 85.96 & 80.70 & 50.00 \\
HermesAgent+DeepSeek-V4-Pro(h) & O1 & 107 & 17.76 & 36.81 & 54.36 & 38.59 \\
HermesAgent+DeepSeek-V4-Pro(h) & O2 & 19 & 15.79 & 42.98 & 21.93 & 62.20 \\
HermesAgent+DeepSeek-V4-Pro(h) & O2$\rightarrow$O1 & 19 & 47.37 & 59.65 & 51.75 & 64.29 \\
HermesAgent+Qwen3.7-Max(m) & O1 & 107 & 23.36 & 43.99 & 59.58 & 32.84 \\
HermesAgent+Qwen3.7-Max(m) & O2 & 19 & 36.84 & 50.88 & 58.77 & 52.44 \\
HermesAgent+Qwen3.7-Max(m) & O2$\rightarrow$O1 & 19 & 52.63 & 64.91 & 64.47 & 41.07 \\
\bottomrule
\end{tabular}
\end{adjustbox}

\caption{RCA results by observability level and observability ablation. RCA-Loc and RCA-ID are task-macro IoUs; Evid. Cov. is pooled golden-action coverage. Scores are percentages. O2$\rightarrow$O1 uses the same 19 originally O2 tasks with regenerated golden actions after direct evidence is made available. Unanswered tasks remain in the denominator. (m)=medium, (h)=high.}
\label{tab:observability_results}
\end{table*}

\begin{table*}[t]
\centering
\small
\setlength{\tabcolsep}{5pt}
\begin{adjustbox}{max width=\textwidth}
\begin{tabular}{lcccc}
\toprule
Agent & Obs. & Gold actions & Agent steps & Agent steps (correct) \\
\midrule
Codex+GPT-5.5(m) & O1 & 6.35 & 32.58 & 32.72 \\
Codex+GPT-5.5(m) & O2 & 4.32 & 37.79 & 31.57 \\
Codex+GPT-5.5(m) & O2$\rightarrow$O1 & 2.95 & 39.58 & 44.43 \\
ClaudeCode+Sonnet-5.5(m) & O1 & 6.35 & 16.08 & 21.87 \\
ClaudeCode+Sonnet-5.5(m) & O2 & 4.32 & 8.37 & 8.00 \\
ClaudeCode+Sonnet-5.5(m) & O2$\rightarrow$O1 & 2.95 & 13.11 & 15.50 \\
HermesAgent+DeepSeek-V4-Pro(h) & O1 & 6.35 & 39.18 & 40.63 \\
HermesAgent+DeepSeek-V4-Pro(h) & O2 & 4.32 & 38.74 & 36.00 \\
HermesAgent+DeepSeek-V4-Pro(h) & O2$\rightarrow$O1 & 2.95 & 34.58 & 30.67 \\
HermesAgent+Qwen3.7-Max(m) & O1 & 6.35 & 28.21 & 24.92 \\
HermesAgent+Qwen3.7-Max(m) & O2 & 4.32 & 24.63 & 25.14 \\
HermesAgent+Qwen3.7-Max(m) & O2$\rightarrow$O1 & 2.95 & 21.79 & 31.20 \\
\bottomrule
\end{tabular}
\end{adjustbox}
\caption{Mean counts per RCA task by observability level. Gold actions are expert-annotated evidence steps; agent steps are distinct executed device--command observations. Agent steps (correct) averages only correctly answered tasks. O1 has 107 tasks; O2 and the O2$\rightarrow$O1 ablation each have 19 tasks. The ablation uses regenerated golden actions. (m)=medium, (h)=high.}
\label{tab:observability_action_counts}
\end{table*}

\subsection{Heterogeneity Results for All Agents}
\label{app:heterogeneity_results}

Table~\ref{tab:heterogeneity_results} reports performance by network heterogeneity level. Heterogeneity captures the need to reason across different device roles, vendors, command syntaxes, and output formats. The stratified results expose differences across heterogeneous settings that are hidden in aggregate performance.

\begin{table*}[t]
\centering
\setlength{\tabcolsep}{5pt}
\begin{adjustbox}{max width=\textwidth}
\begin{tabular}{llcccccc}
\toprule
Agent & Task & H & N & Acc. & Loc. & ID/Rest. & Evid. \\
\midrule
Codex+GPT-5.5(m) & RCA & H-Low ($=1$) & 39 & 66.67 & 71.79 & 71.79 & 48.50 \\
Codex+GPT-5.5(m) & RCA & H-High ($>1$) & 87 & 39.08 & 60.50 & 76.05 & 52.76 \\
Codex+GPT-5.5(m) & Path & H-Low ($\leq6$) & 58 & 89.66 & 100.00 & 94.72 & 49.21 \\
Codex+GPT-5.5(m) & Path & H-High ($>6$) & 50 & 86.00 & 98.67 & 96.22 & 59.39 \\
\midrule
ClaudeCode+Sonnet-5.5(m) & RCA & H-Low ($=1$) & 39 & 56.41 & 71.79 & 83.33 & 16.50 \\
ClaudeCode+Sonnet-5.5(m) & RCA & H-High ($>1$) & 87 & 25.29 & 47.99 & 65.33 & 30.48 \\
ClaudeCode+Sonnet-5.5(m) & Path & H-Low ($\leq6$) & 58 & 18.97 & 98.85 & 67.73 & 8.23 \\
ClaudeCode+Sonnet-5.5(m) & Path & H-High ($>6$) & 50 & 48.00 & 98.00 & 88.73 & 17.91 \\
\midrule
HermesAgent+DeepSeek-V4-Pro(h) & RCA & H-Low ($=1$) & 39 & 28.21 & 43.16 & 39.74 & 29.00 \\
HermesAgent+DeepSeek-V4-Pro(h) & RCA & H-High ($>1$) & 87 & 12.64 & 35.31 & 53.83 & 45.45 \\
HermesAgent+DeepSeek-V4-Pro(h) & Path & H-Low ($\leq6$) & 58 & 44.83 & 91.38 & 74.56 & 24.09 \\
HermesAgent+DeepSeek-V4-Pro(h) & Path & H-High ($>6$) & 50 & 6.00 & 78.67 & 54.81 & 20.13 \\
\midrule
HermesAgent+Qwen3.7-Max(m) & RCA & H-Low ($=1$) & 39 & 46.15 & 56.41 & 55.13 & 31.50 \\
HermesAgent+Qwen3.7-Max(m) & RCA & H-High ($>1$) & 87 & 16.09 & 39.92 & 61.40 & 36.19 \\
HermesAgent+Qwen3.7-Max(m) & Path & H-Low ($\leq6$) & 58 & 48.28 & 91.38 & 73.99 & 20.06 \\
HermesAgent+Qwen3.7-Max(m) & Path & H-High ($>6$) & 50 & 32.00 & 70.00 & 59.92 & 20.36 \\
\bottomrule
\end{tabular}
\end{adjustbox}

\caption{Evaluation results (\%) by network heterogeneity ($H$). Loc. and ID/Rest. are task-macro IoUs for the indicated task type; Evid. is golden-action coverage. (m)=medium, (h)=high.}
\label{tab:heterogeneity_results}
\end{table*}

\subsection{Harness Sensitivity Analysis}
\label{app:harness_sensitivity}

To separate model choice from execution-harness effects, we compare raw
trajectory sets that use the same DeepSeek-V4-Pro model but different agent
harnesses: \texttt{ClaudeCode}, \texttt{HermesAgent}, and \texttt{Codex}. This
analysis treats each harness-model pair as an executable agent because the
harness controls tool scheduling, state serialization, stopping behavior, and
final-answer extraction. All runs use the same task set, gold answers, and
scoring scripts. The comparison is therefore intended as a raw harness
sensitivity analysis under a fixed model family, rather than as an ablation that
isolates each individual harness mechanism. Missing or empty final answers are
kept in the denominator and counted as incorrect.

\begin{table*}[t]
\centering
\scriptsize
\setlength{\tabcolsep}{3pt}
\begin{adjustbox}{max width=\textwidth}
\begin{tabular}{@{}llrrrrrrrr@{}}
\toprule
Task & Harness & Answered & Correct & Acc. & Loc. & ID/Rest. & Path Micro & Evid. F1 & Evid. Cov. \\
\midrule
RCA & ClaudeCode & 124 & 27 & 21.43 & 35.61 & 44.97 & -- & 11.67 & 41.79 \\
RCA & HermesAgent & 124 & 22 & 17.46 & 37.74 & 49.47 & -- & 11.00 & 41.13 \\
RCA & Codex & 125 & 19 & 15.08 & 29.66 & 42.06 & -- & 10.75 & 30.88 \\
Path & ClaudeCode & 108 & 16 & 14.81 & 83.02 & 63.53 & 58.51 & 34.43 & 34.81 \\
Path & HermesAgent & 103 & 29 & 26.85 & 85.49 & 65.42 & 59.26 & 24.06 & 21.52 \\
Path & Codex & 107 & 29 & 26.85 & 82.72 & 65.68 & 59.25 & 28.78 & 24.88 \\
\bottomrule
\end{tabular}
\end{adjustbox}
\caption{Harness sensitivity using DeepSeek-V4-Pro-Preview with high reasoning effort for every row. There are 126 RCA and 108 path-restoration tasks; unanswered tasks remain in the accuracy denominator. Acc., Loc., ID/Rest., Path Micro, and evidence scores are percentages. Loc. and ID/Rest. are task-macro IoUs; ID/Rest. denotes RCA identification or path restoration as appropriate. Path Micro is the pooled directed-edge IoU. Evid. F1 and Evid. Cov. are pooled golden-action F1 and coverage.}
\label{tab:harness_sensitivity_overall}
\end{table*}

For the evidence columns, let $C$ be the total number of covered golden actions,
$G$ the total number of golden actions, and $A$ the sum of per-task counts of
unique executed device commands. We report evidence F1 as
$2C/(G+A)$ and evidence coverage as $C/G$.

\begin{table*}[t]
\centering
\small
\setlength{\tabcolsep}{5pt}
\begin{adjustbox}{max width=\textwidth}
\begin{tabular}{llrrrr}
\toprule
Task & Pair & Both & First only & Second only & Neither \\
\midrule
RCA & CC vs. HA & 15 & 12 & 7 & 92 \\
RCA & CC vs. CX & 16 & 11 & 3 & 96 \\
RCA & HA vs. CX & 12 & 10 & 7 & 97 \\
Path Rest. & CC vs. HA & 6 & 10 & 23 & 69 \\
Path Rest. & CC vs. CX & 10 & 6 & 19 & 73 \\
Path Rest. & HA vs. CX & 11 & 18 & 18 & 61 \\
\bottomrule
\end{tabular}
\end{adjustbox}
\caption{Pairwise exact-answer disagreement under the same DeepSeek-V4-Pro
model. CC, HA, and CX denote ClaudeCode, HermesAgent, and Codex, respectively.
``First only'' and ``Second only'' refer to the order in the Pair column.}
\label{tab:harness_sensitivity_disagreement}
\end{table*}

Table~\ref{tab:harness_sensitivity_overall} shows that CTBench measures a
harness-model combination rather than a model in isolation. On RCA,
ClaudeCode+DeepSeek-V4-Pro(high) obtains the highest exact Accuracy and evidence
coverage among the three harnesses, while HermesAgent+DeepSeek-V4-Pro(high) obtains
the highest macro localization and identification IoU. On Path Restoration,
HermesAgent and Codex reach the same exact Accuracy, HermesAgent has the highest
macro localization IoU, and Codex has the highest macro restoration IoU.
ClaudeCode has the highest path evidence coverage. Comparable resource
measurements for ClaudeCode and Codex are unavailable in this analysis. The
HermesAgent runtime, round, and token values obtained with the unified
resource-accounting method agree with Table~\ref{tab:overall_cost}.

The disagreement counts in Table~\ref{tab:harness_sensitivity_disagreement}
show that different harnesses solve partly different subsets of tasks even when
the underlying model is fixed. For example, on RCA, ClaudeCode solves 11 tasks
that Codex misses, while Codex solves 3 tasks that ClaudeCode misses. On Path
Restoration, Codex solves 19 tasks missed by ClaudeCode, while HermesAgent and
Codex each solve 18 tasks missed by the other.  Overall, these results show that the harness matters even when the underlying model is fixed. Different harnesses solve different subsets of tasks, so we report performance at the harness--model level rather than treating the model as the only source of variation.

\FloatBarrier

\section{Reproducibility Details}
\label{app:reproducibility}

\subsection{Uncertainty Estimation}
\label{app:uncertainty_estimation}

Table~\ref{tab:overall_capability} reports standard errors across tasks. For accuracy and the four task-macro IoUs, we compute the sample standard deviation of the per-task scores divided by the square root of the number of tasks. Evidence coverage is a pooled ratio of covered to total golden actions, rather than a mean of per-task coverage scores. We therefore resample tasks with replacement within each task family (126 RCA or 108 path-restoration tasks), recompute the pooled ratio in each of 10,000 bootstrap samples, and use the standard deviation of those bootstrap estimates as its standard error. The bootstrap uses a fixed base pseudorandom seed of 20260725 with deterministic metric- and model-specific offsets. These seeds govern only the resampling indices and are unrelated to model-generation randomness. Since each task has one Pass@1 execution, these standard errors capture variation across tasks, not variability across repeated runs of the same task.

\subsection{Paired Statistical Comparisons}
\label{app:paired_comparisons}

We compare configurations on the same tasks, retaining unanswered questions under the existing missing-as-wrong scoring rule. Table~\ref{tab:paired_main} compares Codex+GPT-5.5(medium) with the three other configurations in Table~\ref{tab:overall_capability}; Table~\ref{tab:paired_same_harness} compares the two HermesAgent models; and Table~\ref{tab:paired_fixed_model} compares harnesses under the same DeepSeek-V4-Pro-Preview(high) model. Each difference is the estimate for configuration A minus that for B. These comparisons describe harness--model configurations; only the same-harness comparison holds the harness fixed, and only the fixed-model comparison holds the model fixed.

For accuracy, we use the exact two-sided McNemar test on task pairs with discordant correctness and a paired task-level bootstrap interval for the difference in accuracy. For the task-macro localization, identification, and restoration IoUs, we resample task indices jointly for both configurations and recompute the difference in task means. For pooled evidence coverage, each resample recomputes the ratio of total covered to total golden actions for each configuration before taking their difference. The paired bootstrap uses 10,000 resamples and base seed 20260725 with deterministic metric- and model-specific offsets. The intervals are percentile intervals from the paired bootstrap. For IoU and coverage, the exploratory two-sided bootstrap sign probability is twice the smaller fraction of resampled differences on either side of zero, capped at one. We apply Holm correction within each comparison-scope, task-family, and metric family: three comparisons for the main and fixed-model scopes and one for the same-harness scope. The reported 95\% intervals are not multiplicity-adjusted, so an interval excluding zero need not correspond to a Holm-adjusted $p<0.05$. The inference concerns variation across tasks under one Pass@1 execution per task, not run-to-run variation for the same task.

\begin{table*}[t]
\centering
\scriptsize
\setlength{\tabcolsep}{3pt}
\begin{adjustbox}{max width=\textwidth}
\begin{tabular}{@{}lllrrr@{}}
\toprule
Task & Comparator B & Metric & $\Delta$ (pp) & 95\% CI (pp) & $p_{\mathrm{Holm}}$ \\
\midrule
RCA & CC+S & Acc. & +12.70 & $[+3.17, +22.22]$ & 0.0139 \\
RCA & CC+S & Loc. & +8.64 & $[+0.38, +16.96]$ & 0.0400 \\
RCA & CC+S & ID IoU & +3.84 & $[-4.10, +11.51]$ & 0.3348 \\
RCA & CC+S & Evid. Cov. & +24.84 & $[+19.59, +30.03]$ & $<0.001$ \\
RCA & HA+D & Acc. & +30.16 & $[+22.22, +38.89]$ & $<0.001$ \\
RCA & HA+D & Loc. & +26.26 & $[+18.55, +34.01]$ & $<0.001$ \\
RCA & HA+D & ID IoU & +25.26 & $[+17.99, +32.80]$ & $<0.001$ \\
RCA & HA+D & Evid. Cov. & +10.51 & $[+5.44, +15.69]$ & $<0.001$ \\
RCA & HA+Q & Acc. & +22.22 & $[+12.70, +31.75]$ & $<0.001$ \\
RCA & HA+Q & Loc. & +18.97 & $[+11.31, +26.71]$ & $<0.001$ \\
RCA & HA+Q & ID IoU & +15.28 & $[+7.08, +23.88]$ & 0.0016 \\
RCA & HA+Q & Evid. Cov. & +16.69 & $[+11.14, +22.14]$ & $<0.001$ \\
\midrule
Path & CC+S & Acc. & +55.56 & $[+45.37, +65.74]$ & $<0.001$ \\
Path & CC+S & Loc. & +0.93 & $[-1.23, +3.70]$ & 0.5926 \\
Path & CC+S & Rest. IoU & +17.96 & $[+13.95, +21.92]$ & $<0.001$ \\
Path & CC+S & Evid. Cov. & +41.30 & $[+38.04, +44.49]$ & $<0.001$ \\
Path & HA+D & Acc. & +61.11 & $[+51.85, +70.37]$ & $<0.001$ \\
Path & HA+D & Loc. & +13.89 & $[+8.33, +20.06]$ & $<0.001$ \\
Path & HA+D & Rest. IoU & +30.00 & $[+24.59, +35.51]$ & $<0.001$ \\
Path & HA+D & Evid. Cov. & +34.30 & $[+30.37, +37.95]$ & $<0.001$ \\
Path & HA+Q & Acc. & +47.22 & $[+36.11, +57.41]$ & $<0.001$ \\
Path & HA+Q & Loc. & +17.90 & $[+11.42, +25.00]$ & $<0.001$ \\
Path & HA+Q & Rest. IoU & +27.94 & $[+21.24, +34.78]$ & $<0.001$ \\
Path & HA+Q & Evid. Cov. & +35.56 & $[+31.14, +39.71]$ & $<0.001$ \\
\bottomrule
\end{tabular}
\end{adjustbox}
\caption{Paired comparisons of CX+G against each current overall-table configuration. A is CX+G; the reported difference is A minus comparator B. CX+G denotes Codex+GPT-5.5, CC+S ClaudeCode+Sonnet-5.5, HA+D HermesAgent+DeepSeek-V4-Pro, and HA+Q HermesAgent+Qwen3.7-Max. Confidence intervals are unadjusted; $p$ values are Holm-adjusted within each task and metric.}
\label{tab:paired_main}
\end{table*}

\begin{table*}[t]
\centering
\scriptsize
\setlength{\tabcolsep}{3pt}
\begin{adjustbox}{max width=\textwidth}
\begin{tabular}{@{}llrrr@{}}
\toprule
Task & Metric & $\Delta$ (pp) & 95\% CI (pp) & $p_{\mathrm{Holm}}$ \\
\midrule
RCA & Acc. & +7.94 & $[+0.00, +15.87]$ & 0.0755 \\
RCA & Loc. & +7.29 & $[-0.57, +14.93]$ & 0.0656 \\
RCA & ID IoU & +9.99 & $[+1.59, +18.39]$ & 0.0206 \\
RCA & Evid. Cov. & -6.18 & $[-11.23, -1.14]$ & 0.0146 \\
\midrule
Path & Acc. & +13.89 & $[+2.78, +25.00]$ & 0.0275 \\
Path & Loc. & -4.01 & $[-10.19, +2.16]$ & 0.2118 \\
Path & Rest. IoU & +2.06 & $[-4.30, +8.61]$ & 0.5352 \\
Path & Evid. Cov. & -1.26 & $[-5.28, +2.88]$ & 0.5538 \\
\bottomrule
\end{tabular}
\end{adjustbox}
\caption{Paired comparison within HermesAgent: A is HA+Q (Qwen3.7-Max) and B is HA+D (DeepSeek-V4-Pro-Preview). Differences are A minus B in percentage points. Confidence intervals are unadjusted; $p$ values are Holm-adjusted within the single-comparison family for each task and metric.}
\label{tab:paired_same_harness}
\end{table*}

\begin{table*}[t]
\centering
\scriptsize
\setlength{\tabcolsep}{3pt}
\begin{adjustbox}{max width=\textwidth}
\begin{tabular}{@{}lllrrr@{}}
\toprule
Task & A vs. B & Metric & $\Delta$ (pp) & 95\% CI (pp) & $p_{\mathrm{Holm}}$ \\
\midrule
RCA & CC+D vs. CX+D & Acc. & +6.35 & $[+0.79, +11.90]$ & 0.1721 \\
RCA & CC+D vs. CX+D & Loc. & +5.95 & $[-0.67, +12.73]$ & 0.1564 \\
RCA & CC+D vs. CX+D & ID IoU & +2.91 & $[-4.50, +10.45]$ & 0.5748 \\
RCA & CC+D vs. CX+D & Evid. Cov. & +10.91 & $[+6.67, +15.44]$ & $<0.001$ \\
RCA & CC+D vs. HA+D & Acc. & +3.97 & $[-2.38, +11.11]$ & 0.7186 \\
RCA & CC+D vs. HA+D & Loc. & -2.13 & $[-9.75, +5.57]$ & 0.5904 \\
RCA & CC+D vs. HA+D & ID IoU & -4.50 & $[-12.70, +3.84]$ & 0.5748 \\
RCA & CC+D vs. HA+D & Evid. Cov. & +0.66 & $[-4.46, +5.71]$ & 0.8224 \\
RCA & CX+D vs. HA+D & Acc. & -2.38 & $[-8.73, +3.97]$ & 0.7186 \\
RCA & CX+D vs. HA+D & Loc. & -8.08 & $[-15.30, -0.77]$ & 0.0912 \\
RCA & CX+D vs. HA+D & ID IoU & -7.41 & $[-14.81, +0.00]$ & 0.1506 \\
RCA & CX+D vs. HA+D & Evid. Cov. & -10.25 & $[-15.72, -4.70]$ & $<0.001$ \\
\midrule
Path & CC+D vs. CX+D & Acc. & -12.04 & $[-21.30, -3.70]$ & 0.0439 \\
Path & CC+D vs. CX+D & Loc. & +0.31 & $[-5.86, +6.79]$ & 0.9984 \\
Path & CC+D vs. CX+D & Rest. IoU & -2.15 & $[-7.86, +3.31]$ & 1.0000 \\
Path & CC+D vs. CX+D & Evid. Cov. & +9.93 & $[+4.39, +15.42]$ & 0.0004 \\
Path & CC+D vs. HA+D & Acc. & -12.04 & $[-22.22, -1.85]$ & 0.0702 \\
Path & CC+D vs. HA+D & Loc. & -2.47 & $[-8.65, +3.70]$ & 0.9984 \\
Path & CC+D vs. HA+D & Rest. IoU & -1.89 & $[-7.93, +4.24]$ & 1.0000 \\
Path & CC+D vs. HA+D & Evid. Cov. & +13.29 & $[+8.82, +17.73]$ & $<0.001$ \\
Path & CX+D vs. HA+D & Acc. & +0.00 & $[-11.11, +11.11]$ & 1.0000 \\
Path & CX+D vs. HA+D & Loc. & -2.78 & $[-8.64, +3.09]$ & 0.9984 \\
Path & CX+D vs. HA+D & Rest. IoU & +0.26 & $[-6.10, +6.86]$ & 1.0000 \\
Path & CX+D vs. HA+D & Evid. Cov. & +3.36 & $[-1.63, +8.48]$ & 0.1986 \\
\bottomrule
\end{tabular}
\end{adjustbox}
\caption{Paired harness comparisons under fixed DeepSeek-V4-Pro-Preview (high). CC, CX, and HA denote ClaudeCode, Codex, and HermesAgent. Differences are A minus B in percentage points. Confidence intervals are unadjusted; $p$ values are Holm-adjusted across the three harness pairs within each task and metric.}
\label{tab:paired_fixed_model}
\end{table*}

\subsection{Evaluation Environment}
\label{app:evaluation_environment}

All CTBench agent trajectories are executed on a single orchestration host running EulerOS 2.0. The host is equipped with an Intel Xeon Gold 6230N CPU at 2.30GHz, 40 physical cores, 80 logical processors, and 502GB of physical memory. The host runs the agent harnesses, local sandbox workspaces, simulator/tool servers, logging, and result collection. Model inference is not served locally; all evaluated models are accessed through their original vendor APIs.

We pre-collected command-line interface (CLI) logs from the target network devices and served them through a mock backend. For supported commands, the backend returns the recorded device outputs; for invalid queries, it returns the corresponding error messages. For each task, alternative commands that would provide evidence equivalent to a golden diagnostic action are deliberately excluded from the available observation set. Commands that are syntactically valid but absent from the collected logs return a simulated ``permission denied'' response, so agents cannot retrieve the same decisive evidence through an alternative command. This constrained interface supports exact device--command matching as a measure of reference-route coverage, while leaving open whether the agent used the returned observations to form its answer.

We evaluate four harness--LLM configurations with fixed harness versions: Codex (v0.141.0) with GPT-5.5-2026-06-15(medium), ClaudeCode (v2.1.63) with Sonnet-5.5-2026-09-28(medium), and Hermes Agent (v0.16.0) with DeepSeek-V4-Pro-Preview(high) (released April 24, 2026) or Qwen3.7-Max-2026-06-08(medium). To systematically control inference behavior, we explicitly configure the latent reasoning effort (i.e., thinking mode): DeepSeek-V4-Pro is set to the high reasoning level, whereas GPT-5.5, Sonnet-5.5, and Qwen3.7-Max are constrained to the medium level. We did not explicitly set a temperature parameter during evaluation; each harness and model API used its applicable default behavior. Each benchmark task runs in a fresh Docker container. The workspace, tool configuration, logs, and temporary files are recreated for each task to avoid state carryover between runs.

\subsection{Agent Harnesses and Concurrency}
\label{app:harnesses}

We evaluate three harness implementations: \texttt{Codex},
\texttt{ClaudeCode}, and \texttt{HermesAgent}. The reported agent-model
settings are instantiated from these harnesses:
\begin{itemize}
    \item \texttt{Codex+GPT-5.5(medium)},
    \item \texttt{ClaudeCode+Sonnet-5.5(medium)},
    \item \texttt{HermesAgent+DeepSeek-V4-Pro(high)},
    \item \texttt{HermesAgent+Qwen3.7-Max(medium)}.
\end{itemize}
During evaluation, the task queue is capped at three concurrent agent runs to control API request pressure and avoid local simulator contention. Each task has one Pass@1 execution with a 3600-second timeout; we impose no additional limit on the number of steps, interaction rounds, or cumulative tokens per task.

\subsection{Command Form}
\label{app:command_form}

Each harness run follows the same orchestration pattern: select the harness and model, provide the CTBench task JSON, write trajectories and parsed results to a run-specific output directory, and set the maximum concurrency. The model placeholders below refer to the corresponding configurations specified in Section~\ref{app:evaluation_environment}; these commands illustrate the orchestration pattern.

\begin{verbatim}
python <runner>.py --harness codex
  --model <model_name>
  --input <ctbench_task_json>
  --output <result_json>
  --workspace-base <sandbox_root>
  --concurrency 3
  --timeout 3600

python <runner>.py --harness claudecode
  --model <model_name>
  --input <ctbench_task_json>
  --output <result_json>
  --workspace-base <sandbox_root>
  --concurrency 3
  --timeout 3600

python <runner>.py --harness hermesagent
  --model <model_name>
  --input <ctbench_task_json>
  --output <result_json>
  --workspace-base <sandbox_root>
  --concurrency 3
  --timeout 3600
\end{verbatim}

For the Codex harness, the per-task child process created by the runner uses the following command shape, with the task prompt supplied on standard input:

\begin{verbatim}
codex.cmd -a never exec
  --cd <workspace_dir>
  --skip-git-repo-check
  --sandbox workspace-write
  --color never
  --output-last-message last_msg.txt
  --model gpt-5.5
  --ephemeral
\end{verbatim}

\subsection{Prompt Template}
\label{app:prompt_template}

All agent-model combinations use the same task-level prompt template shown in Figure~\ref{fig:prompt-template}. The prompt contains the task description, question identifier, allowed tool interface, and required answer format. It does not include ground-truth answers, task metadata, root-cause categories, difficulty labels, or golden evidence steps.

\begin{figure*}[!htb]
    \centering
    \fbox{%
    \begin{minipage}{0.96\linewidth}
    \textbf{Agent Prompt Template}
    \small

You are an AI agent evaluating comprehensive capabilities for CTBench.

Your task is to solve the given CTBench evaluation problem using only the information and tools allowed by the task.

Follow the task-specific instructions exactly. The task description defines the valid fault categories, answer fields, separators, line order constraints if any, device names, destination formats, and whitespace requirements. These task-specific requirements override any general wording in this prompt.

Work method:
\begin{enumerate}
    \item Identify the affected endpoints, objects, and evaluation scope from the task.
    \item Query only relevant devices and commands needed to confirm or reject plausible root causes.
    \item Prefer direct evidence and cross-checking evidence from the task-allowed command outputs.
    \item Interpret potential faults in the context of the intended network design; do not report a root cause that contradicts the topology, configuration model, or design logic implied by the task.
    \item Do not perform broad exhaustive sweeps when a focused path can answer the task. Avoid repeated equivalent commands after the evidence is sufficient.
    \item If multiple independent root causes exist, return the minimal complete set of root causes. Do not include downstream symptoms or duplicate faults.
    \item If the task restricts the fault type, output only faults of the allowed type.
    \item Use exact device names, interface names, destination IPs, object names, and root-cause labels from the task and observed outputs.
\end{enumerate}

Evidence and safety rules:
\begin{itemize}
    \item Do not use hidden ground truth, answer files, prior trajectory files, scoring scripts, or local simulator output files.
    \item Do not infer answers from filenames, dataset metadata, process state, or filesystem paths.
    \item Query simulator device outputs only through the API or tool interface explicitly allowed by the task.
    \item If an allowed command is unavailable, use the closest allowed command form from the task description; do not treat a missing command output as direct proof of a fault.
\end{itemize}
Before finalizing, silently check:
\begin{itemize}
    \item Does every output line match the exact format requested by the task?
    \item Are all required fields present and in the requested order?
    \item Are separators exactly as requested, usually English semicolons?
    \item Are there no extra spaces, comments, labels, markdown, or explanations?
    \item Are all independent root causes included, and are non-root-cause symptoms excluded?
\end{itemize}
Put only the final answer inside <result></result> tags.

Task:
\{question\_description\}

Question ID:
\{question\_id\}

Return format:
<result>
your final answer here
</result>
    \end{minipage}}
    \caption{Agent Prompt Template.}
    \label{fig:prompt-template}
\end{figure*}

\FloatBarrier

\section{Trajectory-Level RCA Failure Analysis}
\label{app:trajectory_failure_generalization}

\subsection{Overview of Trajectory-Level Failure Analysis}
\label{app:trajectory_failure_overview}

This appendix summarizes a trajectory-level post-hoc analysis of exact-match
RCA failures for \texttt{Codex+GPT-5.5(medium)} and
\texttt{HermesAgent+Qwen3.7-Max(medium)}. The goal is to characterize why an agent's
final answer fails after interacting with the diagnostic environment, rather
than only reporting whether the final answer matches the gold RCA tuple. Among
the 126 RCA tasks, \texttt{Codex+GPT-5.5(medium)} has 66 exact-match failures and
\texttt{HermesAgent+Qwen3.7-Max(medium)} has 94 exact-match failures; 60 tasks are
failed by both systems.

The main failure pattern is not failed tool use. In many cases, agents issue relevant commands and observe useful intermediate evidence, but still fail when mapping that evidence to the final RCA tuple set. The observed errors include missing independent causes, adding non-minimal causes, choosing the wrong fault mechanism, grounding the fault to the wrong node or object, and violating the required output schema.

We identify this pattern through manual inspection of the diagnostic traces, checking each failed task against the task statement, final answer, issued commands, and observed command outputs.

\subsection{Failure Mode Taxonomy}
\label{app:failure_mode_summary}

We organize trajectory failures using the hierarchical RCA failure taxonomy in
Table~\ref{tab:paper_failure_mode_taxonomy}. The taxonomy contains one
output-level failure, two answer-set-level failures, and two entry-level
failures. This taxonomy is used only as a post-hoc explanation of exact-match
errors; it is not a formal component of the benchmark, is not part of the
primary benchmark score, and does not relax the exact-answer criterion. This
structure is important for reproducible annotation: when a prediction both
misses an independent root cause and mislocalizes a returned tuple, the
set-level coverage error is assigned first; entry-level mechanism or grounding
errors are used only when the answer set is otherwise comparable to the gold
set.

For each task, three expert annotators with telecom RCA expertise independently
assigned failure labels, and the taxonomy was applied with a fixed priority
order. Disagreements were resolved through adjudication and discussion under the
same priority rules. The annotators first checked whether the final answer was
parseable under the required RCA schema. If not, the failure was labeled as
answer realization.
For parseable answers, they next compared the predicted and gold root-cause
sets to identify missing independent causes or non-minimal extra causes. Only
after the answer set was comparable did they label entry-level failures:
mechanism discrimination errors when the causal mechanism was wrong, and
fault-object grounding errors when the mechanism was largely correct but the
device, interface, address, policy object, or topology role was incorrect.
Within entry-level annotation, mechanism mismatch has priority over object
mismatch: a prediction with both a wrong mechanism and a wrong object is labeled
F4 rather than F3.

Table~\ref{tab:paper_failure_mode_counts} reports the resulting distribution.
Coverage deficits and mechanism discrimination dominate
\texttt{Codex+GPT-5.5}'s RCA errors, while \texttt{HermesAgent+Qwen3.7-Max}
shows a more distributed error profile with substantially more minimality and
answer-realization failures.

\begin{table*}[!htb]
\centering
\small
\setlength{\tabcolsep}{4pt}
\begin{adjustbox}{max width=\textwidth}
\begin{tabular}{p{0.14\textwidth}p{0.25\textwidth}p{0.15\textwidth}p{0.38\textwidth}}
\toprule
Level & Leaf failure label & Unit & Definition \\
\midrule
Output-level & F5. Answer Realization Failure & Whole answer & The trajectory does not produce a valid final RCA answer, or the answer cannot be parsed into the required tuple schema. \\
Set-level & F1. Causal Coverage Deficit & Answer set & The answer omits one or more independent gold root causes, including dual-root, multi-device, or multi-interface faults. \\
Set-level & F2. Causal Minimality Violation & Answer set & The answer includes redundant, symptom-level, or non-independent causes beyond the minimal gold root-cause set. \\
Entry-level & F4. Mechanism Discrimination Error & Root-cause entry & The returned entry uses the wrong causal mechanism, such as confusing redundancy, routing, policy, address, or forwarding-layer faults. \\
Entry-level & F3. Fault-Object Grounding Error & Root-cause entry & The mechanism is largely correct, but the causal object is grounded to the wrong device, interface, IP address, policy object, underlay/overlay layer, or active/standby role. \\
\bottomrule
\end{tabular}
\end{adjustbox}
\caption{Hierarchical RCA failure taxonomy used for trajectory-level error analysis. Rows follow the annotation priority order.}
\label{tab:paper_failure_mode_taxonomy}
\end{table*}

\begin{table}[!htb]
\centering
\small
\setlength{\tabcolsep}{4pt}
\begin{adjustbox}{max width=\textwidth}
\begin{tabular}{lrrrrrr}
\toprule
Model & F1 & F2 & F3 & F4 & F5 & Total \\
\midrule
\texttt{GPT-5.5} & 28 & 1 & 13 & 23 & 1 & 66 \\
\texttt{Qwen3.7-Max} & 32 & 16 & 18 & 16 & 12 & 94 \\
\bottomrule
\end{tabular}
\end{adjustbox}
\caption{Failure-mode counts for exact-match RCA failures. F1--F5 refer to the
labels in Table~\ref{tab:paper_failure_mode_taxonomy}; each failed trajectory
is assigned one primary failure label.}
\label{tab:paper_failure_mode_counts}
\end{table}

\subsection{Representative Failure Trajectories}
\label{app:representative_trajectory_examples}

The full trajectories for these examples are provided in the Code and Data
Supplement under \texttt{model\_trajectory}, organized by model name, task type,
and task identifier.
All representative examples below were checked against the recorded final
answers and diagnostic traces of the corresponding agents.

\paragraph{F5 Answer realization failure: RCA q12.}
The gold answer is
\texttt{FW\_02;114.114.114.114;global VRRP hot standby redundancy protocol not enabled}.
\texttt{HermesAgent+Qwen3.7-Max} returns an empty final answer despite successfully
interacting with the environment. The trajectory lacks valid result tags, separators,
or parseable lines, leading to a complete output failure independent of the agent's
actual domain knowledge.

\paragraph{F1 Causal coverage deficit: RCA q7.}
The gold answer contains two independent root causes:
\texttt{FW\_01;10.3.10.1;security policy rule does not permit the corresponding user}
and
\texttt{BJHQ\_CSR1000V\_GW\_01;10.1.120.251;IP prefix list missing the corresponding user source IP address}.
\texttt{Codex+GPT-5.5} outputs only the firewall security-policy root cause and
omits the prefix-list fault, prematurely halting its diagnosis once the first
locally plausible cause is identified.

\paragraph{F2 Causal minimality violation: RCA q18.}
The gold answer contains a VRRP redundancy configuration fault on
\texttt{FW\_02}, while \texttt{HermesAgent+Qwen3.7-Max} returns both a
prefix-list fault and a VRRP fault on a different firewall. Although this answer
also contains localization errors, the set-level F2 label takes precedence because
the final submitted set retains unpruned, non-gold causal assertions alongside downstream symptoms.

\paragraph{F4 Mechanism discrimination error: RCA q10.}
The gold root cause is
\texttt{FW\_02;114.114.114.114;global VRRP hot standby redundancy protocol not enabled}.
\texttt{Codex+GPT-5.5} localizes the answer to the correct firewall and address
but incorrectly reports an OSPF configuration error, successfully grounding the location
while confusing redundancy-state semantics with routing-protocol configurations.

\paragraph{F3 Fault-object grounding error: RCA q11.}
The gold root cause is
\texttt{FW\_02;8.8.8.8;global VRRP hot standby redundancy protocol not enabled}.
\texttt{HermesAgent+Qwen3.7-Max} outputs
\texttt{Core\_SW\_01;8.8.8.8;global VRRP hot standby redundancy protocol not enabled}.
The mechanism and address are aligned with the gold answer, but the causal
device is misattributed to a neighboring core switch rather than the faulty firewall itself.
The agent fails to decouple its observation point from the actual causal node.

\subsection{Implications for Agent Design}
\label{app:agent_design_implications}

The failure taxonomy suggests several checks that are useful for troubleshooting agents. During a run, the agent should track which symptoms have been explained, which candidate causes remain possible, which observations support or reject each candidate, and whether the final answer satisfies the required schema.

The observed failure modes suggest several useful checks for agent design:
\begin{itemize}
    \item \textbf{Causal Coverage (F1) \& Minimality (F2):} A global diagnostic state tracker may help ensure all symptoms are explained without including redundant downstream effects.
    \item \textbf{Semantic Grounding (F3 \& F4):} Domain-specific verification may reduce mechanism confusion and mislocalization by checking protocol constraints (e.g., routing vs. redundancy states) and maintaining role-aware topology mapping to separate observation nodes from causal origins.
    \item \textbf{Robust Finalization (F5):} Deterministic output validation can enforce schema constraints before final submission.
\end{itemize}

Overall, this fine-grained analysis suggests that, in these RCA failures, current agents more often struggle with multi-step diagnostic reasoning and causal minimization than with basic tool execution.

\FloatBarrier
\section{Discussion of Agent Capabilities in CTBench}
\label{app:discussion_beyond_telecom_rca}

Although CTBench is grounded in telecom network operations, the trajectory-level
RCA failures reflect broader agent capabilities required in complex professional
environments. The goal of this discussion is to clarify how the failure modes
summarized in Table~\ref{tab:paper_failure_mode_taxonomy} relate to general
agent capabilities and how these pressures arise from CTBench task
characteristics. In many failed trajectories, agents can collect relevant
observations but still fail to maintain causal coverage, prune downstream
symptoms, ground evidence to the correct object, distinguish similar mechanisms,
or realize the final answer in the required structured format.

These failure modes are consistent with the task properties of CTBench RCA:
heterogeneous device roles, partial observability, layered protocol mechanisms,
causal propagation, multiple independent faults, and multi-step evidence
requirements. Table~\ref{tab:paper_ctbench_generalization} summarizes how the
observed failure modes relate to broader agent capabilities and the CTBench
properties that stress them.

\begin{table*}[htb]
\centering
\small
\setlength{\tabcolsep}{4pt}
\begin{adjustbox}{max width=\textwidth}
\begin{tabular}{p{0.28\textwidth}p{0.34\textwidth}p{0.30\textwidth}}
\toprule
Failure taxonomy & General agent capability & CTBench property \\
\midrule
F1 Causal coverage deficit & Causal attribution and long-horizon evidence planning & \texttt{root\_cause\_count}; \texttt{golden\_solution\_path} \\
F2 Causal minimality violation & Causal attribution and symptom pruning & \texttt{fault\_propagation\_chain}; \texttt{golden\_solution\_path} \\
F3 Fault-object grounding error & Interaction with heterogeneous environments & \texttt{network\_heterogeneity} \\
F4 Mechanism discrimination error & Partial-observation reasoning and causal attribution & \texttt{protocol\_complexity}; \texttt{evidence\_observability} \\
F5 Answer realization failure & Reliable structured answer realization & Output schema constraints \\
\bottomrule
\end{tabular}
\end{adjustbox}
\caption{Qualitative mapping from CTBench RCA failure modes to broader agent capabilities and related task properties.}
\label{tab:paper_ctbench_generalization}
\end{table*}

\end{document}